\documentclass[twoside,11pt]{article}

\usepackage{jmlr2e}

\usepackage{booktabs}
\usepackage{subfigure}
\usepackage{soul}
\usepackage{color}
\usepackage{amsmath, amsfonts}
\usepackage{mymacros}
\usepackage{algorithm}
\usepackage{algpseudocode}
\usepackage{algorithmicx}
\usepackage{url}
\usepackage{microtype}

\newcommand{\trans}{^\top}
\newcommand{\transs}{^{~\top}}

\jmlrheading{1}{2000}{1-48}{4/00}{10/00}{Benigno Uria, Marc-Alexandre C\^ot\'e, Karol Gregor, Iain Murray, Hugo Larochelle}

\ShortHeadings{Neural Autoregressive Distribution Estimation}{Uria, C\^ot\'e, Gregor, Murray, Larochelle}
\firstpageno{1}

\begin{document}

\title{Neural Autoregressive Distribution Estimation}

\author{\name Benigno Uria \email benigno.uria@gmail.com\\
       \addr Google DeepMind\\
       London, UK
       \AND
       \name Marc-Alexandre C\^ot\'e \email marc-alexandre.cote@usherbrooke.ca \\
       \addr Department of Computer Science\\
       Universit\'e de Sherbrooke\\
       Sherbrooke, QC, Canada
       \AND
       \name Karol Gregor \email karol.gregor@gmail.com\\
       \addr Google DeepMind\\
       London, UK
       \AND
       \name Iain Murray \email i.murray@ed.ac.uk\\
       \addr School of Informatics\\
       University of Edinburgh\\
       Edinburgh, UK
       \AND
       \name Hugo Larochelle \email hlarochelle@twitter.com \\
       \addr Twitter Cortex\\
       Cambridge, MA, USA}

\editor{Leslie Pack Kaelbling}

\maketitle

\begin{abstract}
We present Neural Autoregressive Distribution Estimation (NADE) models,
which are neural network architectures applied to the problem of
unsupervised distribution and density estimation. They leverage
the probability product rule and a weight sharing scheme inspired
from restricted Boltzmann machines, to yield an estimator that
is both tractable and has good generalization performance.
We discuss how they achieve competitive performance in modeling both binary and
real-valued observations. We also present how deep NADE models can
be trained to be agnostic to the ordering of input dimensions used
by the autoregressive product rule decomposition. Finally, we also show
how to exploit the topological
structure of pixels in images using a deep convolutional architecture for NADE\@.
\end{abstract}


\section{Introduction}

Distribution estimation is one of the most general problems addressed by machine learning.
From a good and flexible distribution estimator, in principle it is
possible to solve a variety of types of inference problem,
such as classification, regression, missing value imputation, and
many other predictive tasks.

Currently, one of the most common forms of distribution estimation is
based on directed graphical models. In general these models describe
the data generation process as sampling a latent state $\vec{h}$
from some prior $p(\vec{h})$, followed by sampling the observed data
$\vec{x}$ from some conditional $p(\vec{x}\g\vec{h})$. Unfortunately,
this approach quickly becomes intractable and requires approximations
when the latent state $\vec{h}$ increases in complexity. Specifically,
computing the marginal probability of the data,
$p(\vec{x}) = \sum_{\vec{h}} p(\vec{x}\g\vec{h})\, p(\vec{h})$, is only
tractable under fairly constraining assumptions on
$p(\vec{x}\g\vec{h})$ and $p(\vec{h})$.

Another popular approach, based on undirected graphical models, gives
probabilities of the form $p(\vec{x}) = \exp{\phi(\vec{x})} / Z$,
where $\phi$ is a tractable function and $Z$ is a normalizing constant. A popular choice for such
a model is the restricted Boltzmann machine (RBM), which
substantially out-performs
mixture models on a variety of binary datasets \citep{Salakhutdinov2008}.
Unfortunately, we often cannot compute probabilities $p(\vec{x})$ exactly in
undirected models either, due to the normalizing constant~$Z$.

In this paper, we advocate a third approach to distribution
estimation, based on autoregressive models and feed-forward neural networks. We
refer to our particular approach as Neural Autoregressive Distribution Estimation (NADE)\@. Its main
distinguishing property is that computing $p(\vec{x})$ under a NADE model
is tractable and can be computed efficiently, given an arbitrary ordering of
the dimensions of $\vec{x}$. We show that the framework is flexible and
can model both binary and real-valued observations,
can be made order-agnostic, and can be adapted to the case of 2D images
using convolutional neural networks. In each case, we're able to reach
competitive results, compared to popular directed and undirected
graphical model alternatives.

\section{NADE}
\label{sec:NADE}

We consider the problem of modeling the distribution $p(\vec{x})$ of
input vector observations $\vec{x}$. For now, we will assume that the
dimensions of $\vec{x}$ are binary, that is $x_d \in \{0,1\}~\forall d$.
The model generalizes to other data types, which is explored later
(Section~\ref{sec:nonbinary}) and in other work (Section~\ref{sec:conclusion}).

NADE begins with the observation
that any $D$-dimensional distribution $p(\vec{x})$ can be factored into a product of one-dimensional distributions, in any order $o$ (a permutation of the integers $1,\dots,D$):
\begin{equation}
p(\vec{x}) = \prod_{d=1}^D p(x_{o_d} \given \xoltd) \mpdot
\label{eq:autoregressive}
\end{equation}
Here $\oltd$ contains the first $d-1$
dimensions in ordering $o$ and $\xoltd$ is the corresponding subvector
for these dimensions. Thus, one can define an `autoregressive' generative
model of the data simply by specifying a parameterization of all $D$ conditionals
$p(\xod \given \xoltd)$.

\citet{FreyB1996} followed this approach and proposed using simple (log-)linear
logistic regression models for these conditionals. This choice yields surprisingly
competitive results, but are not competitive with non-linear models such as an RBM\@.
\citet{Bengio2000} proposed a more flexible approach, with
a single-layer feed-forward neural network for each conditional. Moreover,
they allowed connections between the output of each network and the hidden layer of
networks for the conditionals appearing earlier in the autoregressive ordering.
Using neural networks led to some improvements in modeling performance, though
at the cost of a really large model for very high-dimensional data.

In NADE, we also model each conditional using a feed-forward neural network.
Specifically, each conditional $p(x_{o_d}\given \xltd)$ is parameterized as follows:
\begin{align}
p(x_{o_d} \te 1 \given
\xoltd) & = \sigm{\mat{V}_{o_d,\cdot}\vec{h}_d+b_{o_d}}\label{eq:nade-output}\\
\vec{h}_d & = \sigm{\mat{W}_{\cdot,o_{<d}}\xoltd + \vec{c}}\mpcomma
\label{eq:hidden-units}
\end{align}
where $\sigm{a} = 1/(1+e^{-a})$ is the logistic sigmoid, and with $H$ as the
number of hidden units, $\mat{V} \in \realdomain^{D\times H}$, $\vec{b} \in
\realdomain^D$, $\mat{W} \in \realdomain^{H \times D }$, $\vec{c} \in
\realdomain^H$ are the parameters of the NADE model.

\begin{figure}
    \centering
    \includegraphics[width=0.8\textwidth]{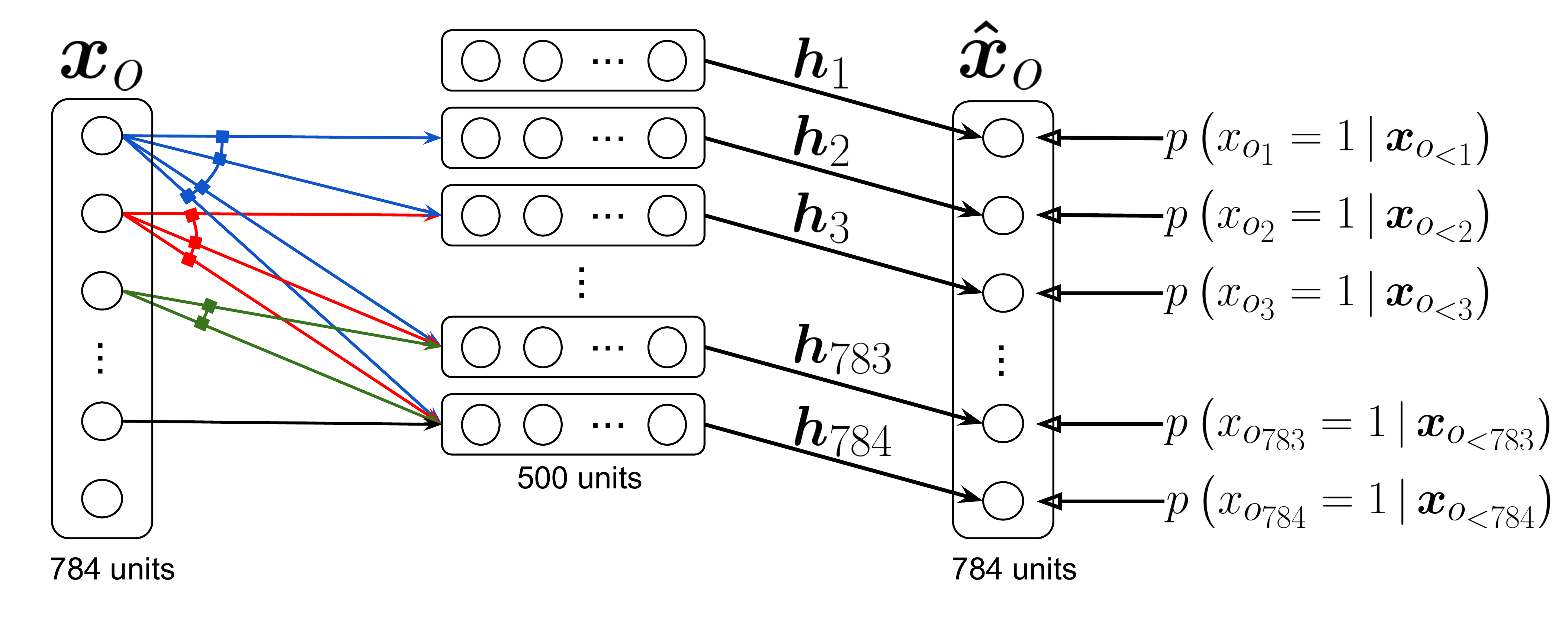}
    \caption[NADE]{
      Illustration of a NADE model. In this example, in the input layer, units with value 0 are shown in black while units with value 1 are shown in white. The dashed border represents a layer pre-activation.The outputs $\hat\x_O$ give predictive probabilities for each dimension of a vector $\x_O$, given elements earlier in some ordering. There is no path of connections between an output and the value being predicted, or elements of $\x_O$ later in the ordering. Arrows connected together correspond to connections with shared (tied) parameters.
    }
    \label{fig:nade}
  \end{figure}

The hidden layer matrix $\mat{W}$ and bias $\vec{c}$
are shared by each hidden layer $\vec{h}_d$ (which are all of the
same size). This parameter sharing scheme (illustrated in Figure~\ref{fig:nade}) means that NADE has $O(HD)$ parameters, rather than $O(HD^2)$ required if the neural networks were separate. Limiting the number of parameters can reduce the risk of over-fitting. Another advantage is that all $D$ hidden
layers $\vec{h}_d$ can be computed in $O(HD)$ time instead of $O(HD^2)$. Denoting
the pre-activation of the $d^{\rm th}$ hidden layer as
$\vec{a}_d = \mat{W}_{\cdot,o_{<d}}\xoltd + \vec{c}$, this complexity
is achieved by using the recurrence
\begin{align}
\vec{h}_1 = &\sigm{\vec{a}_1}, \quad\text{where~} \vec{a}_1 = \vec{c}\\
\vec{h}_d = &\sigm{\vec{a}_d}, \quad\text{where~} \vec{a}_d = \mat{W}_{\cdot,o_{<d}}\xoltd + \vec{c} = \mat{W}_{\cdot,o_{d-1}} x_{o_{d-1}} + \vec{a}_{d-1}\label{eq:rec-hidden-units}\\
&\mbox{\hspace*{9cm}~~~for $d\in\{2,\dots,D\}$}\mpcomma \nonumber
\end{align}
where Equation~\ref{eq:rec-hidden-units} given vector $\vec{a}_{d-1}$ can be computed in $O(H)$.
Moreover, the computation of Equation~\ref{eq:nade-output} given $\vec{h}$ is also $O(H)$.
Thus, computing $p(\vec{x})$ from $D$ conditional distributions (Equation~\ref{eq:autoregressive})
costs $O(HD)$ for NADE\@. This complexity is comparable to that of regular feed-forward neural network
models.

NADE can be trained by maximum likelihood, or equivalently by minimizing the average negative log-likelihood,
\begin{align}
  \frac{1}{N}\sum_{n=1}^N -\log p(\vec{x}^{(n)})& = \frac{1}{N}\sum_{n=1}^N \sum_{d=1}^D -\log p(x_{o_d}^{(n)}\g\xoltd^{(n)})\mpcomma
\end{align}
usually by stochastic (minibatch) gradient descent. As probabilities $p(\vec{x})$ cost $O(HD)$, gradients
of the negative log-probability of training examples can also be computed in $O(HD)$. Algorithm~\ref{alg:nade} describes
the computation of both $p(\vec{x})$ and the gradients of $-\log p(\vec{x})$ with respect to NADE's parameters.

\begin{algorithm}[tb]
  \caption{Computation of $p(\vec{x})$ and learning gradients for NADE\@.}
  \label{alg:nade}
  \begin{algorithmic}
    \State {\bfseries Input:} training observation vector $\vec{x}$ and ordering $o$ of the input dimensions.
    \State {\bfseries Output:} $p(\vec{x})$ and gradients of $-\log p(\vec{x})$ on parameters.
    \State
    \State \# Computing $p(\vec{x})$
    \State $\vec{a}_1 \leftarrow \vec{c}$
    \State $p(\vec{x}) \leftarrow 1$
    \For{$d$ from 1 to $D$}
    \State $\vec{h}_d \leftarrow \sigm{\vec{a}_d}$
    \State $p(x_{o_d}\te1\g\xoltd) \leftarrow \sigm{\mat{V}_{o_d,\cdot}\vec{h}_d+b_{o_d}}$
    \State $p(\vec{x}) \leftarrow p(\vec{x}) \left(p(x_{o_d}\te1\g\xoltd)^{x_{o_d}}+(1-p(x_{o_d}\te1\g\xoltd))^{1-x_{o_d}}\right)$
    \State $\vec{a}_{d+1} \leftarrow \vec{a}_d + \mat{W}_{\cdot,o_d} x_{o_d}$
    \EndFor
    \State
    \State \# Computing gradients of $-\log p(\vec{x})$
    \State $\delta \vec{a}_D \leftarrow 0$
    \State $\delta \vec{c} \leftarrow 0$
    \For{$d$ from $D$ to 1}
    \State $\delta b_{o_d} \leftarrow \left(p(x_{o_d}\te1\g\xoltd) - x_{o_d}\right)$
    \State $\delta \mat{V}_{o_d,\cdot} \leftarrow \left(p(x_{o_d}\te1\g\xoltd) - x_{o_d}\right) \vec{h}_d\trans$
    \State $\delta \vec{h}_d \leftarrow \left(p(x_{o_d}\te1\g\xoltd) - x_{o_d}\right) \mat{V}_{o_d,\cdot}\transs$
    \State $\delta \vec{c} \leftarrow \delta \vec{c} + \delta \vec{h}_d\odot \vec{h}_d \odot (1-\vec{h}_d)$
    \State $\delta \mat{W}_{\cdot,o_d} \leftarrow \delta \vec{a}_{d} x_{o_d}$
    \State $\delta \vec{a}_{d-1} \leftarrow \delta \vec{a}_d + \delta \vec{h}_d\odot \vec{h}_d \odot (1-\vec{h}_d)$
    \EndFor
    \State
    \Return $p(\vec{x}), \delta \vec{b}, \delta \mat{V}, \delta \vec{c}, \delta \mat{W}$
  \end{algorithmic}
\end{algorithm}

\subsection{Relationship with the RBM}
\label{sec:NADE-and-RBM}

The proposed weight-tying for NADE isn't simply motivated by computational reasons.
It also reflects the computations of approximation inference in the RBM\@.

Denoting the energy function and distribution under an RBM as
\begin{align}
  E(\vec{x},\vec{h}) &= -\vec{h}\trans \mat{W} \vec{x} -\vec{b}\trans \vec{x} - \vec{c}\trans \vec{h}\\
  p(\vec{x},\vec{h}) &= \exp{-E(\vec{x},\vec{h})} / Z\,,
\end{align}
computing all conditionals
\begin{align}
  p(\vec{x}_{o_d}\g\xoltd) &= \sum_{\vec{x}_{o_{>d}}\in \{0,1\}^{D-d}} \;\sum_{\vec{h}\in \{0,1\}^{H}} \exp{-E(\vec{x},\vec{h})} / Z(\xoltd)\\
  Z(\vec{\xoltd}) &= \sum_{\vec{x}_{o_{\geq d}}\in \{0,1\}^{D-d+1}} \;\sum_{\vec{h}\in \{0,1\}^{H}} \exp{-E(\vec{x},\vec{h})}   \,
\end{align}
is intractable. However, these could be approximated using mean-field variational inference. Specifically, consider the conditional over $x_{o_d}$, $\vec{x}_{o_{>d}}$ and $\vec{h}$ instead:
\begin{align}
  p(x_{o_d},\vec{x}_{o_{>d}}, \vec{h}\g\xoltd) &= \exp{-E(\vec{x},\vec{h})} / Z(\xoltd)\mpdot
\end{align}
A mean-field approach could first approximate this conditional with a factorized distribution
\begin{align}
q(x_{o_d},\vec{x}_{o_{>d}},\vec{h}\g\xoltd) =\; & \mu_i(d)^{x_{o_d}} (1-\mu_d(d))^{1-x_{o_d}} \prod_{j>d} \mu_j(d)^{x_{o_j}} (1-\mu_j(d))^{1-x_{o_j}} \nonumber\\
                                   & \prod_k \tau_k(d)^{h_k} (1-\tau_k(d))^{1-h_k}\mpcomma \label{eqn:mf-family}
\end{align}
where $\mu_j(d)$ is the marginal probability of $x_{o_j}$
being equal to 1, given $\xoltd$.  Similarly, $\tau_k(d)$ is the
marginal for hidden variable $h_k$. The
dependence on $d$ comes from conditioning on $\xoltd$, that is on the first $d\!-\!1$ dimensions
of $\vec{x}$ in ordering~$o$.

For some $d$, a mean-field approximation is obtained
by finding the parameters $\mu_j(d)$ for $j\in\{d,\dots,D\}$ and $\tau_k(d)$ for $k\in\{1,\dots,H\}$ which
minimize the KL divergence between $q(x_{o_d},\vec{x}_{o_{>d}},\vec{h}\g\xoltd)$
and $p(x_{o_d},\vec{x}_{o_{>d}},\vec{h}\g\xoltd)$. This is usually done by finding message passing updates
that each set the derivatives of the KL
divergence to 0 for some of the parameters of $q(x_{o_d},\vec{x}_{o_{>d}},\vec{h}\g\xoltd)$ given others.

For some $d$, let us fix $\mu_j(d)=x_{o_d}$ for $j<d$, leaving only $\mu_j(d)$ for $j>d$ to be
found. The KL-divergence develops as follows:
\begin{eqnarray*}
   & & {\rm KL}(q(x_{o_d},\vec{x}_{o_{>d}},\vec{h}\g\vec{x}_{o_{<d}}) \;||\; p(x_{o_d},\vec{x}_{o_{>d}},\vec{h}\g\vec{x}_{o_{<d}}) )\\
   & = & - \sum_{x_{o_d},\vec{x}_{o_{>d}},\vec{h}} q(x_{o_d},\vec{x}_{o_{>d}},\vec{h}\g\vec{x}_{o_{<d}}) \log p(x_{o_d},\vec{x}_{o_{>d}},\vec{h}\g\vec{x}_{o_{<d}}) \\
   &   & + \sum_{x_{o_d},\vec{x}_{o_{>d}},\vec{h}} q(x_{o_d},\vec{x}_{o_{>d}},\vec{h}\g\vec{x}_{o_{<d}}) \log q(x_{o_d},\vec{x}_{o_{>d}},\vec{h}\g\vec{x}_{o_{<d}}) \\
   & = & \log Z(\vec{x}_{o_{<d}}) - \sum_j \sum_k \tau_k(d) W_{k,o_j} \mu_j(d) - \sum_j b_{o_j} \mu_j(d) - \sum_k c_k \tau_k(d) \\
   &   & + \sum_{j\geq d} \left( \mu_j(d) \log \mu_j(d) + (1-\mu_j(d)) \log (1-\mu_j(d))\right) \\
   &   & + \sum_{k} \left( \tau_k(d) \log \tau_k(d) + (1-\tau_k(d)) \log (1-\tau_k(d))\right)\mpdot
\end{eqnarray*}

Then, we can take the derivative with respect to $\tau_k(d)$ and set it to 0, to obtain:
\begin{eqnarray}
   0 & = & \frac{\partial {\rm KL}(q(x_{o_d},\vec{x}_{o_{>d}},\vec{h}\g\vec{x}_{o_{<d}}) \;||\; p(x_{o_d},\vec{x}_{o_{>d}},\vec{h}\g\vec{x}_{o_{<d}}) )}{\partial \tau_k(d)} \nonumber\\
   0 & = & - c_k -  \sum_j W_{k,o_j} \mu_j(d)  + \log \left(\frac{\tau_k(d)}{1-\tau_k(d)}\right)\nonumber\\
   \frac{\tau_k(d)}{1-\tau_k(d)} & = & \exp{c_k + \sum_j W_{k,o_j} \mu_j(d)}\\
   \tau_k(d) & = & \frac{\exp{c_k + \sum_j W_{k,o_j} \mu_j(d)}}{1+\exp{c_k + \sum_j W_{k,o_j} \mu_j(d)}}\nonumber\\
   \tau_k(d) & = & \sigm{c_k + \sum_{j\geq d} W_{k,o_j} \mu_j(d) + \sum_{j < d} W_{k,o_j} x_{o_j} }\mpdot \label{eq:mf-hidden}
\end{eqnarray}
where in the last step we have used the fact that $\mu_j(d) = x_{o_j}$ for $j<d$. Equation~\ref{eq:mf-hidden} would
correspond to the message passing updates of the hidden unit marginals $\tau_k(d)$ given
the marginals of input $\mu_j(d)$.

Similarly, we can set the derivative with respect to $\mu_j(d)$ for $j\geq d$ to 0 and obtain:
\begin{eqnarray}
   0 & = & \frac{\partial {\rm KL}(q(x_{o_d},\vec{x}_{o_{>d}},\vec{h}\g\vec{x}_{o_{<d}}) \;||\; p(x_{o_d},\vec{x}_{o_{>d}},\vec{h}\g\vec{x}_{o_{<d}}) )}{\partial \mu_j(d)} \nonumber\\
   0 & = & - b_{o_d} - \sum_k \tau_k(d) W_{k,o_j}   + \log \left(\frac{\mu_j(d)}{1-\mu_j(d)}\right)\nonumber\\
   \frac{\mu_j(d)}{1-\mu_j(d)} & = & \exp{ b_{o_j} + \sum_k \tau_k(d) W_{k,o_j} }\nonumber\\
   \mu_j(d) & = & \frac{\exp{b_{o_j} + \sum_k \tau_k(d) W_{k,o_j} }}{1+\exp{b_{o_j} + \sum_k \tau_k(d) W_{k,o_j} }}\nonumber\\
   \mu_j(d) & = & \sigm{b_{o_j} + \sum_{k} \tau_k(d) W_{k,o_j} }\mpdot \label{eq:mf-input}
\end{eqnarray}
Equation~\ref{eq:mf-input} would correspond to the message passing updates of the input marginals $\mu_j(d)$
given the hidden layer marginals $\tau_k(d)$. The complete mean-field algorithm would thus alternate between applying the
updates of Equations~\ref{eq:mf-hidden}~and~\ref{eq:mf-input}, right to left.

We now notice that Equation~\ref{eq:mf-hidden} corresponds to NADE's hidden layer computation (Equation~\ref{eq:hidden-units}) where $\mu_j(d) = 0~\;\forall j\geq d$.
Also, Equation~\ref{eq:mf-input} corresponds to NADE's output layer computation (Equation~\ref{eq:nade-output}) where $j=d$, $\tau_k(d)=h_{d,k}$ and $\mat{W}\trans = \mat{V}$.
Thus, in short, NADE's forward pass is equivalent to applying a single pass of mean-field inference to
approximate all the conditionals $p(\vec{x}_{o_d}\g\xoltd)$ of an RBM, where initially $\mu_j(d) = 0$ and where a separate matrix $\mat{V}$ is used
for the hidden-to-input messages. A generalization of NADE based on this connection to mean field inference has been further explored by \citet{raiko2014}.

\section{NADE for non-binary observations}
\label{sec:nonbinary}

So far we have only considered the case of binary observations $x_i$. However, the
framework of NADE naturally extends to distributions over other types of
observations.

In the next section, we discuss the case of real-valued observations, which is
one of the most general cases of non-binary observations and provides an
illustrative example of the technical considerations one faces when extending
NADE to new observations.

\subsection{RNADE: Real-valued NADE}
A NADE model for real-valued data could be obtained by applying the derivations shown in Section~\ref{sec:NADE-and-RBM} to the Gaussian-RBM~\citep{Welling2005}. The resulting neural network would output the mean of a Gaussian with fixed variance for each of the conditionals in Equation~\ref{eq:autoregressive}. Such a model is not competitive with mixture models, for example on perceptual datasets~\citep{Uria2015b}. However, we can explore alternative models by making the neural network for each conditional distribution output the parameters of a distribution that's not a fixed-variance Gaussian.

In particular, a mixture of one-dimensional Gaussians for each autoregressive conditional provides a flexible model. Given enough components, a mixture of Gaussians can model any continuous distribution to arbitrary precision. The resulting model can be interpreted as a sequence of mixture density networks~\citep{Bishop1994} with shared parameters.
We call this model RNADE-MoG\@. In RNADE-MoG, each of the conditionals is modeled by a mixture of Gaussians:
\begin{align}
p(\xod \given \xoltd) = &
\sum_{c=1}^{C} \pi_{\od,c} \;\gaussian(\xod;\, \mu_{\od,c},\, \sigma_{\od,c}^2)\mpcomma
\end{align}
where the parameters are set by the outputs of a neural network:
\begin{align}
\pi_{\od,c} = &
\frac{\exp{z_{\od,c}\pp{\pi}}}{\sum_{c=1}^{C}\exp{z_{\od,c}\pp{\pi}}}
\\
\mu_{\od,c} = &
z_{\od,c}\pp{\mu}
\\
\sigma_{\od,c} = &
\exp{z_{\od,c}\pp{\sigma}}
\\
z_{\od,c}\pp{\pi} = &
b_{\od,c}\pp{\pi} + \sum_{k=1}^{H} V_{\od,k,c}\pp{\pi} h_{d,k}
\label{eq:RNADE-outputs-pi}
\\
z_{\od,c}\pp{\mu} = &
b_{\od,c}\pp{\mu} + \sum_{k=1}^{H} V_{\od,k,c}\pp{\mu} h_{d,k}
\label{eq:RNADE-outputs-mu}
\\
z_{\od,c}\pp{\sigma} = &
b_{\od,c}\pp{\sigma} + \sum_{k=1}^{H} V_{\od,k,c}\pp{\sigma} h_{d,k}
\label{eq:RNADE-outputs-sigma}
\end{align}

Parameter sharing conveys the same computational and statistical advantages as it does in the binary NADE\@.

Different one dimensional conditional forms may be preferred, for example due to limited dataset size or domain knowledge about the form of the conditional distributions. Other choices, like single variable-variance Gaussians, sinh-arcsinh distributions, and  mixtures of Laplace distributions, have been examined by~\citet{Uria2015b}.

Training an RNADE can still be done by stochastic gradient descent on the parameters of the model with respect to the negative log-density of the training set. It was found empirically~\citep{Uria2013} that stochastic gradient descent leads to better parameter configurations when the gradient of the mean $\left(\frac{\partial J}{\partial\mu_{\od,c}}\right)$ was multiplied by the standard deviation ($\sigma_{\od,c}$).

\section{Orderless and Deep NADE}
\label{sec:DeepNADE}

The fixed ordering of the variables in a NADE model makes the exact calculation of arbitrary conditional probabilities computationally intractable. Only a small subset of conditional distributions, those where the conditioned variables are at the beginning of the ordering and marginalized variables at the end, are computationally tractable.

Another limitation of NADE is that a naive extension to a deep version, with multiple layers of hidden units, is computationally expensive. Deep neural networks~\citep{Bengio2009, LeCun2015} are at the core of state-of-the-art models for supervised tasks like image recognition~\citep{Krizhevsky2012} and speech recognition~\citep{Dahl2013}. The same inductive bias should also provide better unsupervised models. However, extending the NADE framework to network architectures with several hidden layers, by introducing extra non-linear calculations between Equations~\eqref{eq:hidden-units} and~\eqref{eq:nade-output}, increases its complexity to cubic in the number of units per layer. Specifically, the cost becomes $O(DH^2L)$, where $L$ stands for the number of hidden layers and can be assumed to be a small constant, $D$ is the number of variables modeled, and $H$ is the number of hidden units, which we assumed to be of the same order as $D$. This increase in complexity is caused by no longer being able to share hidden layer computations across the conditionals in Equation~\ref{eq:autoregressive}, after the non-linearity in the first layer.

In this section we introduce an order-agnostic training procedure, DeepNADE, which will address both of the issues above. This procedure trains a single deep neural network that can assign a conditional distribution to any variable given any subset of the others. This network can then provide the conditionals in Equation~\ref{eq:autoregressive} for any ordering of the input observations. Therefore, the network defines a factorial number of different models with shared parameters, one for each of the $D!$ orderings of the inputs. At test time, given an inference task, the most convenient ordering of variables can be used. The models for different orderings will not be consistent with each other: they will assign different probabilities to a given test vector. However, we can use the models' differences to our advantage by creating ensembles of NADE models (Section~\ref{sec:nade-ensembles}), which results in better estimators than any single NADE\@. Moreover, the training complexity of our procedure increases linearly with the number of hidden layers $O(H^2L)$, while remaining quadratic in the size of the network's layers.

We first describe the model for an $L$-layer neural network modeling binary variables. A conditional distribution is obtained directly from a hidden unit in the final layer:
\begin{align}
    p(\xod\te1 \given \xoltd, \params, \oltd, \od) &= \vec{h}^{(L)}_{\od} 
\mpdot
\label{eq:deepnade:output}
\end{align}
This hidden unit is computed from previous layers, all of which can only
depend on the $\xoltd$ variables that are currently being conditioned on. We
remove the other variables from the computation using a binary mask,
\begin{align}
\moltd &= [1_{1\in \oltd}, 1_{2\in \oltd}, \dots, 1_{D\in \oltd}]\mpcomma
\label{eq:deepnade:mask}
\end{align}
which is element-wise multiplied with the inputs before computing the remaining
layers as in a standard neural network:
\begin{align}
\vec{h}^{(0)} &=
\x \odot \moltd \label{eq:deepnade:input}\\
\vec{a}^{(\ell)} &=
\mat{W}^{(\ell)} \vec{h}^{(\ell-1)} + \vec{b}^{(\ell)} \label{eq:deepnade:hidden-preactivation}\\
\vec{h}^{(\ell)} &=
\bsigma{\vec{a}^{(\ell)}} \label{eq:deepnade:hidden}\\
\vec{h}^{(L)} &=
\bsigm{\vec{a}^{(L)}} \label{eq:deepnade:last-hidden}\mpdot
\end{align}
The network is specified by a free choice of the activation function $\bsigma{\cdot}$, and
learnable parameters $\mat{W}^{(\ell)} \in \realdomain^{H^{(\ell)}
\times H^{(\ell-1)}}$ and $\vec{b}^{(\ell)} \in \realdomain^{H^{(\ell)}}$, where
$H^{(l)}$ is the number of units in the $\ell$-th layer. As layer zero is the
masked input, $H^{(0)}=D$. The final $L$-th
layer needs to be able to provide predictions for any element
(Equation~\ref{eq:deepnade:output}) and so also has $D$ units.

To train a DeepNADE, the ordering of the variables is treated as a stochastic variable with a uniform distribution. Moreover, since we wish DeepNADE to provide good predictions for any ordering, we optimize the expected likelihood over the ordering of variables:
\begin{align}
\loss(\params) & = \expectationorderings -\log p(\vec{X}
\given \params,\order) \propto
\expectationorderings \expectationtrainingdata -\log
p(\vx \given \params, o)\mpcomma \label{eq:loss-function}
\end{align}
where we've made the dependence on the ordering $o$ and the network's parameters $\params$ explicit, $D!$ stands for the set of all orderings (the permutations of $D$ elements) and $\vx$ is a uniformly sampled datapoint from the training set $\cal{X}$. Using NADE's expression for the density of a datapoint in Equation~\ref{eq:autoregressive} we have
\begin{equation}
\loss(\params)=\expectationorderings \expectationtrainingdata
\sum_{d=1}^D -\log p(\xod \given \xoltd, \params, o)
\mpcomma
\end{equation}
where $d$ indexes the elements in the ordering, $o$, of the variables. By moving the expectation over orderings inside the sum over the elements of the ordering, the ordering can be split in three parts: $\oltd$ (the indices of the $d - 1$ first dimensions in the ordering), $\od$ (the index of the $d$-th variable) and $\ogtd$ (the indices of the remaining dimensions). Therefore, the loss function can be rewritten as:
\begin{equation}
\loss(\params)= \expectationtrainingdata\sum_{d=1}^D
\expectationoltd \expectationod \expectationogtd
- \log p(\xod \given \xoltd, \params, \oltd, \od, \ogtd)
\mpdot
\end{equation}
The value of each of these terms does not depend on $o_{>d}$. Therefore, it can be simplified as:
\begin{equation}
\loss(\params)= \expectationtrainingdata\sum_{d=1}^D
\expectationoltd \expectationod
- \log p(\xod \given \xoltd, \params, \oltd, \od)
\mpdot
\label{eq:loss-intractable}
\end{equation}

In practice, this loss function will have a very
high number of terms and will have to be approximated by sampling $\vx$, $d$ and $o_{<d}$. The innermost expectation over values of $o_d$ can be calculated cheaply, because all of the neural network computations depend only on the masked input $\xoltd$, and can be reused for each possible $o_d$. Assuming all orderings are equally probable, we will estimate $\loss(\params)$ by:
\begin{align}
\widehat{\loss}(\params) &= \dfrac{D}{D-d+1}\sum_{o_d} -\log
p(\xod \given \xoltd, \params, \oltd, \od )
\mpcomma
\label{eq:hatloss}
\end{align}
which is an unbiased estimator of Equation~\ref{eq:loss-function}. Therefore, training
can be done by descent on the gradient of $\widehat{\loss}(\params)$.

For binary observations, we use the cross-entropy scaled by a factor of $\frac{D}{D-d+1}$ as the training loss which corresponds to minimizing $\widehat{\loss}$:
\begin{align}
\mathcal{J}(\vec{x}) &=
  \frac{D}{D-d+1} \; \mogted\trans \left(\vec{x} \odot \blog{\vec{h}^{(L)}} + (1-\vec{x}) \odot \blog{1-\vec{h}^{(L)}} \right)
  \mpdot \label{eq:deepnade:binary:loss}
\end{align}
Differentiating this cost involves backpropagating the gradients of the cross-entropy only from the outputs in $o_{\geq d}$ and rescaling them by $\frac{D}{D-d+1}$.

The resulting training procedure resembles that of a denoising autoencoder \citep{vincent2008}.
Like the autoencoder, $D$ outputs are used to predict $D$ inputs corrupted by a random masking process ($\moltd$ in Equation~\ref{eq:deepnade:input}).
A single forward pass can compute $\vec{h}^{(L)}_{o_{\geq d}}$, which provides a prediction $p(\xod\te1 \given \xoltd, \params, o_{<d}, o_d)$ for every masked variable, which could be used next in an ordering starting with $o_{<d}$. Unlike the autoencoder, the outputs for variables corresponding to those provided in the input (not masked out) are ignored.

\begin{figure}
    \centering
    \includegraphics[width=\textwidth]{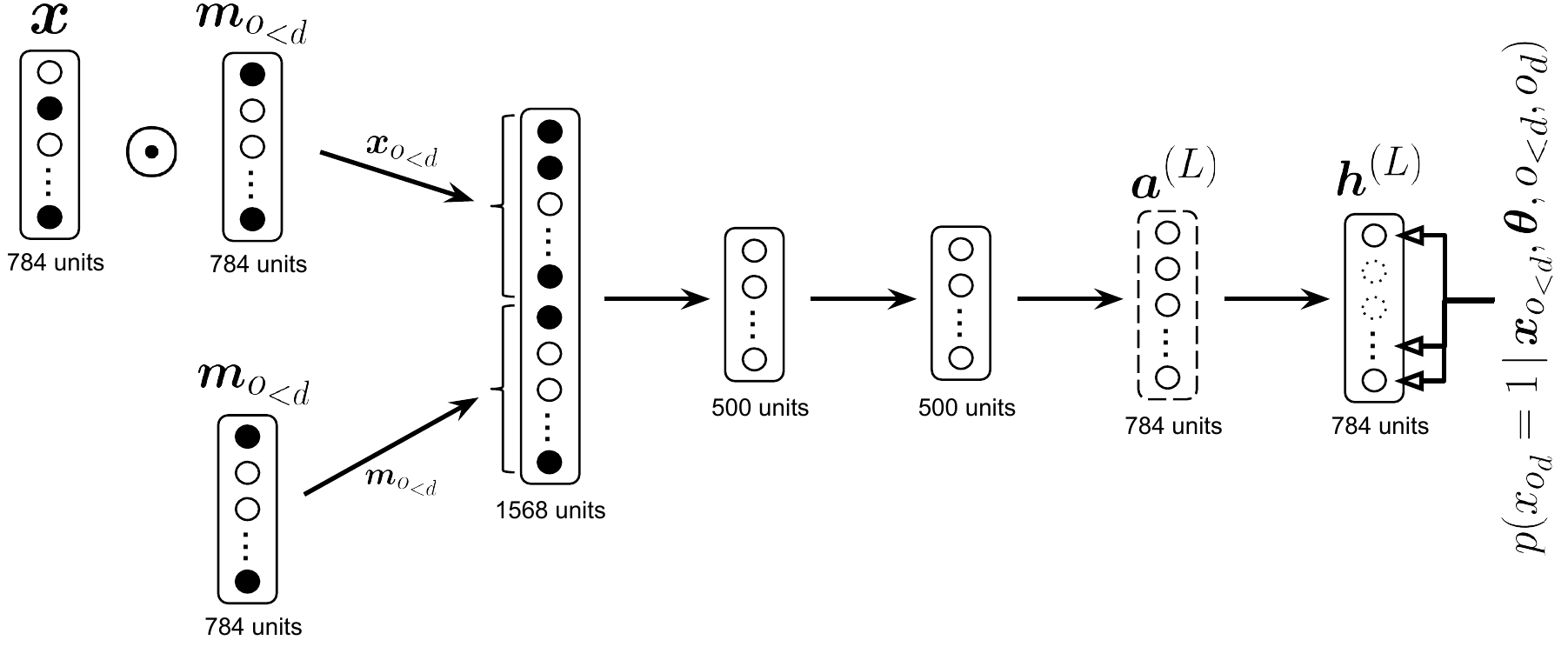}
    \caption[ConvNADE]{
      Illustration of a DeepNADE model with two hidden layers. The dashed border represents a layer pre-activation. A mask $\moltd$ specifies a subset of variables to condition on. A conditional or predictive probability of the remaining variables is given in the final layer.
      Note that the output units with a corresponding input mask of value 1 (shown with dotted contour) are effectively not involved in DeepNADE's training loss (Equation~\ref{eq:deepnade:binary:loss}). 
    }
    \label{fig:deepnade}
  \end{figure}

In this order-agnostic framework, missing variables and zero-valued observations are indistinguishable by the network. This shortcoming can be alleviated by concatenating the inputs to the network (masked variables $\vx \odot \moltd$) with the mask $\moltd$. Therefore we advise substituting the input described in Equation~\ref{eq:deepnade:input} with
\begin{align}
\vec{h}^{(0)} &= \concat(\x \odot \moltd, \moltd) \mpdot \label{eq:deepnade_input_and_mask}
\end{align}
We found this modification to be important in order to obtain competitive statistical performance (see Table~\ref{tab:deepnade-mnist-results}). The resulting neural network is illustrated in Figure~\ref{fig:deepnade}.

\subsection{Ensembles of NADE models}
\label{sec:nade-ensembles}
As mentioned, the DeepNADE parameter fitting procedure effectively produces a factorial number of different NADE models, one for each ordering of the variables. These models will not, in general, assign the same probability to any particular datapoint. This disagreement is undesirable if we require consistent inferences for different inference problems, as it will preclude the use of the most convenient ordering of variables for each inference task.

However, it is possible to use this variability across the different orderings
to our advantage by combining several models. A usual approach to improve on
a particular estimator is to construct an ensemble of multiple, strong but
different estimators, e.g.\ using bagging~\citep{Ormoneit1995} or
stacking~\citep{Smyth1999}. The DeepNADE training procedure suggests a way
of generating ensembles of NADE models: take a set of uniformly distributed
orderings $\{o^{(k)}\}_{k=1}^K$ over the input variables and use the average
probability $\frac{1}{K} \sum_{k=1}^K p(\vec{x} \g \theta, o^{(k)})$ as an
estimator.

The use of an ensemble increases the test-time cost of density estimation linearly with the number of orderings used. The complexity of sampling does not change however: after one of the $K$ orderings is chosen at random, the single corresponding NADE is sampled. Importantly, the cost of training also remains the same, unlike other ensemble methods such as bagging. Furthermore, the number of components can be chosen after training and even adapted to a computational budget on the fly.

\section{ConvNADE: Convolutional NADE}
  \label{sect:convnade}

  One drawback of NADE (and its variants so far) is the lack of a mechanism for truly exploiting the high-dimensional structure of the data. For example, when using NADE on binarized MNIST, we first need to flatten the 2D images before providing them to the model as a vector. As the spatial topology is not provided to the network, it can't use this information to share parameters and may learn less quickly.

  Recently, convolutional neural networks (CNN) have achieved state-of-the-art performance on many supervised tasks related to images~\cite{Krizhevsky2012}. Briefly, CNNs are composed of convolutional layers, each one having multiple learnable filters. The outputs of a convolutional layer are feature maps and are obtained by the convolution on the input image (or previous feature maps) of a linear filter, followed by the addition of a bias and the application of a non-linear activation function. Thanks to the convolution, spatial structure in the input is preserved and can be exploited. Moreover, as per the definition of a convolution the same filter is reused across all sub-regions of the entire image (or previous feature maps), yielding a parameter sharing that is natural and sensible for images.

  The success of CNNs raises the question: can we exploit the spatial topology of the inputs while keeping NADE's autoregressive property? It turns out we can, simply by replacing the fully connected hidden layers of a DeepNADE model with convolutional layers. We thus refer to this variant as Convolutional NADE (ConvNADE)\@.

  First we establish some notation that we will use throughout this section. Without loss of generality, let the input $\mat{X} \in \binarydomain^{N_X \times N_X}$ be a square binary image of size $N_X$ and every convolution filter $\mat{W}^{(\ell)}_{ij} \in \realdomain^{N_W^{(\ell)} \times N_W^{(\ell)}}$ connecting two feature maps $\mat{H}^{(\ell-1)}_i$ and $\mat{H}^{(\ell)}_j$ also be square with their size $N_W^{(\ell)}$ varying for each layer $\ell$. We also define the following mask $\Moltd \in \binarydomain^{N_X \times N_X}$, which is 1 for the locations of the first $d-1$ pixels in the ordering $o$.

  Formally, Equation~\ref{eq:deepnade:hidden-preactivation} is modified to use convolutions instead of dot products. Specifically for an $L$-layer convolutional neural network that preserves the input shape (explained below) we have
  \begin{align}
    p(\xod\te1 \given \xoltd, \params, o_{<d}, o_d) &=
      \bvec{\mat{H}^{(L)}_1}_{o_d} \mpcomma \label{eq:convnade:output}
  \end{align}
  with
  \begin{align}
    \mat{H}^{(0)}_1 &=
      \X \odot \Moltd \label{eq:convnade:input}\\
    \mat{A}^{(\ell)}_j &= b^{(\ell)}_j +
      \sum_{i=1}^{H^{(\ell-1)}} \mat{H}^{(\ell-1)}_i \circledast \mat{W}^{(\ell)}_{ij} \label{eq:convnade:hidden-preactivation}\\
    \mat{H}^{(\ell)}_j &=
      \bsigma{\mat{A}^{(\ell)}_j} \label{eq:convnade:hidden}\\
    \mat{H}^{(L)}_j &=
      \bsigm{\mat{A}^{(L)}_j} \label{eq:convnade:last-hidden}
    \mpcomma
  \end{align}
  where $H^{(\ell)}$ is the number of feature maps output by the $\ell$-th layer and $\vec{b}^{(l)} \in \realdomain^{H^{(l)}}$, $\mat{W}^{(\ell)} \in \realdomain^{H^{(\ell-1)} \times H^{(\ell)} \times N_W^{(\ell)} \times N_W^{(\ell)}}$, with $\odot$ denoting the element-wise multiplication, $\bsigma{\cdot}$ being any activation function and $\bvec{\mat{X}}\rightarrow\x$ is the concatenation of every row in $\mat{X}$. Note that $H^{(0)}$ corresponds to the number of channels the input images have.

  For notational convenience, we use $\circledast$ to denote both ``valid'' convolutions and ``full'' convolutions, instead of introducing bulky notations to differentiate these cases. The ``valid'' convolutions only apply a filter to complete patches of the image, resulting in a smaller image (its shape is decreased to $N_X - N_W^{(\ell)} + 1$). Alternatively, ``full'' convolutions zero-pad the contour of the image before applying the convolution, thus expanding the image (its shape is increased to $N_X + N_W^{(\ell)} - 1$). Which one is used should be self-explanatory depending on the context. Note that we only use convolutions with a stride of 1.

  Moreover, in order for ConvNADE to output conditional probabilities as shown in Equation~\ref{eq:convnade:output}, the output layer must have only one feature map $\mat{H}^{(L)}_1$, whose dimension matches the dimension of the input $\X$. This can be achieved by carefully combining layers that use either ``valid'' or ``full'' convolutions.

  To explore different model architectures respecting that constraint, we opted for the following strategy. Given a network, we ensured the first half of its layers was using ``valid'' convolutions while the other half would use ``full'' convolutions. In addition to that, we made sure the network was symmetric with respect to its filter shapes (\ie the filter shape used in layer $\ell$ matched the one used in layer $L-\ell$).

  For completeness, we wish to mention that ConvNADE can also include pooling and upsampling layers, but we did not see much improvement when using them. In fact, recent research suggests that these types of layers are not essential to obtain state-of-the-art results~\citep{Springenberg2015}.

  The flexibility of DeepNADE allows us to easily combine both convolutional and fully connected layers. To create such hybrid models, we used the simple strategy of having two separate networks, with their last layer fused together at the end. The `convnet' part is only composed of convolutional layers whereas the `fullnet' part is only composed of fully connected layers. The forward pass of both networks follows respectively Equations~\eqref{eq:convnade:input}--\eqref{eq:convnade:hidden} and Equations~\eqref{eq:deepnade:input}--\eqref{eq:deepnade:hidden}. Note that in the `fullnet' network case, $\x$ corresponds to the input image having been flattened.

  In the end, the output layer $\vec{g}$ of the hybrid model corresponds to the aggregation of the last layer pre-activation of both `convnet' and `fullnet' networks. The conditionals are slightly modified as follows:
  \begin{align}
    p(\xod\te1 \given \xoltd, \params, o_{<d}, o_d) &=
      \vec{g}_{o_d} \label{eq:convnade-hybrid:output}\\
    \vec{g} &=
      \bsigm{\bvec{\mat{A}_1^{(L)}} + \vec{a}^{(L)}} \label{eq:convnade-hybrid:last-hidden}
    \mpdot
  \end{align}

  The same training procedure as for DeepNADE model can also be used for ConvNADE\@. For binary observations, the training loss is similar to Equation~\ref{eq:deepnade:binary:loss}, with $\vec{h}^{(L)}$ being substituted for $\vec{g}$ as defined in Equation~\ref{eq:convnade-hybrid:last-hidden}.

  As for the DeepNADE model, we found that providing the mask $\Moltd$ as an input to the model improves performance (see Table~\ref{tab:mnist-2d-results}). For the `convnet' part, the mask was provided as an additional channel to the input layer. For the `fullnet' part, the inputs were concatenated with the mask as shown in Equation~\ref{eq:deepnade_input_and_mask}.

  \begin{figure}
    \centering
    \includegraphics[width=\textwidth]{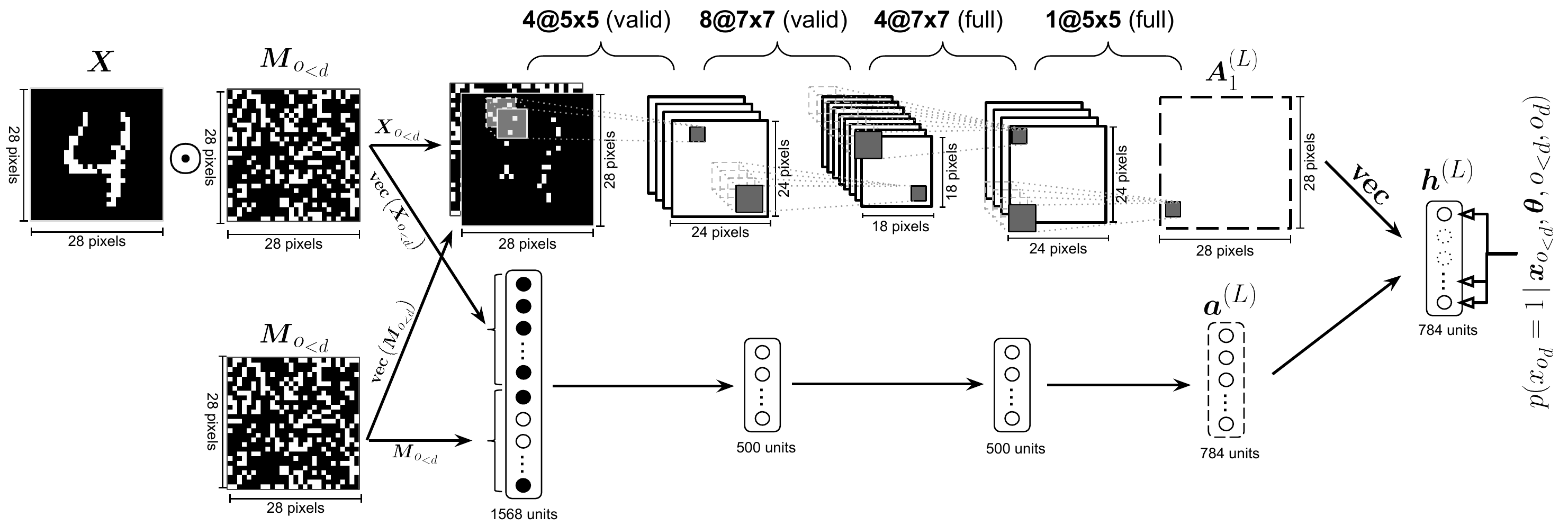}
    \caption[ConvNADE]{
      Illustration of a ConvNADE that combines a convolutional neural network with three hidden layers and a fully connected feed-forward neural network with two hidden layers. The dashed border represents a layer pre-activation. Units with a dotted contour are not valid conditionals since they depend on themselves \ie they were given in the input.
    }
    \label{fig:convnade}
  \end{figure}

  The final architecture is shown in Figure~\ref{fig:convnade}. In our experiments, we found that this type of hybrid model works better than only using convolutional layers (see Table~\ref{tab:mnist-2d-results}). Certainly, more complex architectures could be employed but this is a topic left for future work.

\section{Related Work}

As we mentioned earlier, the development of NADE and its extensions was motivated
by the question of whether a tractable distribution estimator could be designed
to match a powerful but intractable model such as the restricted Boltzmann machine.

The original inspiration came from the autoregressive approach taken by
fully visible sigmoid belief networks (FVSBN), which were shown by \citet{FreyB1996}
to be surprisingly competitive, despite the simplicity of the distribution
family for its conditionals. \citet{Bengio2000} later proposed using
more powerful conditionals, modeled as single layer neural networks. Moreover, they
proposed connecting the output of each $d^{\rm th}$ conditional to all of the hidden
layers of the $d-1$ neural networks for the preceding conditionals. More recently,
\citet{Germain2015} generalized this model by deriving a simple procedure for
making it deep and orderless (akin to DeepNADE, in Section~\ref{sec:DeepNADE}).
We compare with all of these approaches in Section~\ref{sec:binvec}.

There exists, of course, more classical and non-autoregressive approaches to tractable distribution estimation,
such as mixture models and Chow--Liu trees~\citep{chow1968}. We compare with these as well in Section~\ref{sec:binvec}.

This work also relates directly to the recently growing literature on generative neural networks.
In addition to the autoregressive approach described in this paper, there exists three other types
of such models: directed generative networks, undirected generative networks and hybrid networks.

Work on directed generative networks dates back to the original work on sigmoid belief networks~\citep{Neal92}
and the Helmholtz machine~\citep{Hinton95,Dayan95}. Helmholtz machines are equivalent to a multilayer sigmoid
belief network, with each using binary stochastic units. Originally they were trained using Gibbs sampling and gradient descent \citep{Neal92},
or with the so-called wake sleep algorithm \citep{Hinton95}. More recently, many
alternative directed models and training procedures have been proposed. \citet{KingmaD2014, RezendeD2014} proposed the variational
autoencoder (VAE), where the model is the same as the Helmholtz machine, but with real-valued (usually Gaussian) stochastic units. Importantly, \citet{KingmaD2014} identified a reparameterization trick making it possible to
train the VAE in a way that resembles the training of an autoencoder. This approach falls in the family of stochastic variational
inference methods, where the encoder network corresponds to the approximate variational posterior.
The VAE optimizes a bound on the likelihood which is estimated using a single sample from the variational posterior, though recent work
has shown that a better bound can be obtained using an importance sampling approach~\citep{BurdaY2016}.
\citet{GregorK2015} later exploited the VAE approach to develop DRAW, a directed generative model for images
based on a read-write attentional mechanism. \citet{GoodfellowI2014} also proposed an adversarial approach to training
directed generative networks, that relies on a discriminator network simultaneously trained to distinguish
between data and model samples. Generative networks trained this way are referred to as Generative Adversarial
Networks (GAN)\@. While the VAE optimizes a bound of the likelihood (which is the KL divergence
between the empirical and model distributions), it can
be shown that GAN optimizes the Jensen--Shannon (JS) divergence between the empirical and model distributions.
\citet{LiY2015} instead propose a training objective derived from Maximum Mean Discrepancy \citep[MMD;][]{GrettonA2007}.
Recently, the directed generative model approach has been very successfully applied to model images~\citep{DentonE2015,SohlJ2011}.

The undirected paradigm has also been explored extensively for developing powerful generative networks.
These include the restricted Boltzmann machine~\citep{Smolensky86,Freund+Haussler92} and its multilayer extension,
the deep Boltzmann machine~\citep{Salakhutdinov+Hinton-2009}, which dominate the literature on undirected neural networks.
\citet{Salakhutdinov2008} provided one of the first quantitative evidence of the generative modeling
power of RBMs, which motivated the original parameterization for NADE~\citep{Larochelle2011}.
Efforts to train better undirected models can vary in nature. One has been to develop
alternative objectives to maximum likelihood. The proposal of Contrastive Divergence \citep[CD;][]{Hinton2002} was instrumental in
the popularization of the RBM\@. Other proposals include pseudo-likelihood~\citep{Besag75pseudolikelihood,Marlin10InductivePrinciples},
score matching~\citep{Hyvarinen-2005,Hyvarinen-2007,Hyvarinen-2007b}, noise contrastive estimation~\citep{Gutmann+Hyvarinen-2010}
and probability flow minimization~\citep{SohlJ2011}. Another line of development has been to optimize likelihood using Robbins–Monro
stochastic approximation~\citep{YounesL1989}, also known as Persistent CD~\citep{Tieleman08}, and
develop good MCMC samplers for deep undirected models~\citep{SalakhutdinovR2009,Salakhutdinov-ICML2010,Desjardins+al-2010,Cho10IJCNN}.
Work has also been directed towards proposing improved update rules or
parameterization of the model's energy function~\citep{TielemanT2009,ChoK2013,MontavonG2012} as
well as improved approximate inference of the hidden layers~\citep{Salakhutdinov+Larochelle-2010}.
The work of \citet{NgiamJ2011} also proposed an undirected model that
distinguishes itself from deep Boltzmann machines by having deterministic hidden units, instead of stochastic.

Finally, hybrids of directed and undirected networks are also possible, though much less common. The most notable
case is the Deep Belief Network \citep[DBN;][]{Hinton06}, which corresponds to a sigmoid belief network for
which the prior over its top hidden layer is an RBM (whose hidden layer counts as an additional hidden layer).
The DBN revived interest in RBMs, as they were required to successfully initialize the DBN\@.

NADE thus substantially differs from this literature focusing on directed and undirected models,
benefiting from a few properties that these approaches lack. Mainly, NADE does not rely
on latent stochastic hidden units, making it possible to tractably compute its associated
data likelihood for some given ordering. This in turn makes it possible to efficiently produce
exact samples from the model (unlike in undirected models) and get an unbiased
gradient for maximum likelihood training (unlike in directed graphical models).

\section{Results}
  \label{sec:results}
  In this section, we evaluate the performance of our different NADE models on a variety of datasets.

  \subsection{Binary vectors datasets}
    \label{sec:binvec}

    We start by evaluating the performance of NADE models on a set of benchmark datasets where the observations
    correspond to binary vectors. These datasets were mostly taken from the LIBSVM datasets web
    site\footnote{\url{http://www.csie.ntu.edu.tw/~cjlin/libsvmtools/datasets/}},
    except for OCR-letters\footnote{\url{http://ai.stanford.edu/~btaskar/ocr/}} and NIPS-0-12\,\footnote{\url{http://www.cs.nyu.edu/~roweis/data.html}}.
    Code to download these datasets is available here: \url{http://info.usherbrooke.ca/hlarochelle/code/nade.tar.gz}.
    Table~\ref{tab:uci_datasets_info} summarizes the main statistics for these datasets.

    \begin{table}
    \caption{Statistics on the binary vector datasets of Section~\ref{sec:binvec}.}
    \begin{center}
    \begin{tabular}{lrrrr}\toprule
    Name & \# Inputs & Train & Valid\rlap{.} & Test \\
    \midrule
    Adult       & 123 &  5000 &  1414 &  26147 \\
    Connect4    & 126 & 16000 &  4000 &  47557 \\
    DNA         & 180 &  1400 &   600 &   1186 \\
    Mushrooms   & 112 &  2000 &   500 &   5624 \\
    NIPS-0-12   & 500 &   400 &   100 &   1240 \\
    OCR-letters & 128 & 32152 & 10000 &  10000 \\
    RCV1        & 150 & 40000 & 10000 & 150000 \\
    Web         & 300 & 14000 &  3188 &  32561 \\
    \bottomrule
    \label{tab:uci_datasets_info}
    \end{tabular}
    \end{center}
    \end{table}

    For these experiments, we only consider tractable distribution estimators,
    where we can evaluate $p(\x)$ on test items exactly. We consider the following baselines:
    \begin{itemize}
    \item {\bf MoB}: A mixture of multivariate Bernoullis, trained using
      the EM algorithm. The number of mixture components was chosen from $\{32, 64, 128, 256, 512,
      1024\}$ based on validation set performance, and early stopping was
      used to determine the number of EM iterations.
    \item {\bf RBM}: A restricted Boltzmann machine made tractable by
      using only 23 hidden units, trained by contrastive divergence with up to 25 steps
      of Gibbs sampling. The validation set performance was used to select the learning rate
      from $\{0.005, 0.0005, 0.00005\}$, and the number
      of iterations over the training set from $\{100, 500, 1000\}$.
    \item {\bf FVSBN}: Fully visible sigmoid belief network, that models each
      conditional $p(x_{o_d}\g \xoltd)$ with logistic regression. The ordering
      of inputs was selected randomly. Training was by stochastic gradient
      descent. The validation set was used for
      early stopping, as well as for choosing the base learning rate $\eta \in \{0.05, 0.005, 0.0005\}$,
      and a decreasing schedule constant $\gamma$ from $\{0,0.001,0.000001\}$ for the
      learning rate schedule $\eta / (1+\gamma t)$ for the $t^{\rm th}$ update.
    \item {\bf Chow--Liu}:
      A Chow--Liu tree is a graph over the observed variables, where the distribution of each variable,
      except the root, depends on a single parent node. There is an
      $O(D^2)$ fitting algorithm to find the maximum likelihood tree and
      conditional distributions \citep{chow1968}.
      We adapted an implementation provided by \citet{harmeling2011}, who found
      Chow--Liu to be a strong baseline.

      The maximum likelihood parameters are not defined when conditioning
      on events that haven't occurred in the training set. Moreover, conditional
      probabilities of zero are possible, which could give infinitely bad test set
      performance. We re-estimated the conditional probabilities on the Chow--Liu
      tree using Lidstone or ``add-$\alpha$'' smoothing:
      \begin{equation}
          p(x_d\te1 \given x_{\mathrm{parent}}\te z) = \frac{\mathrm{count}(x_d\te1\given x_{\mathrm{parent}}\te z) + \alpha}{\mathrm{count}(x_{\mathrm{parent}}\te z) + 2\alpha}\mpcomma
      \end{equation}
      selecting $\alpha$ for each dataset from $\{10^{-20}, 0.001, 0.01, 0.1\}$
      based on performance on the validation set.

    \item {\bf MADE}~\citep{Germain2015}: Generalization of the neural network approach of \citet{Bengio2000}, to multiple
      layers. We consider a version using a single (fixed) input ordering and another trained
      on multiple orderings from which an ensemble was constructed (which was inspired from
      the order-agnostic approach of Section~\ref{sec:DeepNADE}) that we refer to as MADE-E\@. See \citet{Germain2015} for more details.
    \end{itemize}

    We compare these baselines with the two following NADE variants:
    \begin{itemize}
    \item {\bf NADE (fixed order)}: Single layer NADE model, trained on a single (fixed) randomly generated order, as described in
    Section~\ref{sec:NADE}. The sigmoid activation function was used for the hidden layer, of size 500. Much like for FVSBN, training relied
    on stochastic gradient descent and the validation set was used for
    early stopping, as well as for choosing the learning rate from \{0.05, 0.005, 0.0005\},
      and the decreasing schedule constant $\gamma$ from \{0,0.001,0.000001\}.
    \item {\bf NADE-E}: Single layer NADE trained according to the order-agnostic procedure described in Section~\ref{sec:DeepNADE}.
    The rectified linear activation function was used for the hidden layer, also of size 500. Minibatch gradient descent was used for training,
    with minibatches of size 100. The initial learning rate, chosen among \{0.016, 0.004, 0.001, 0.00025, 0.0000675\}, was linearly
    decayed to zero over the course of $100,000$ parameter updates. Early stopping was used, using
    Equation~\ref{eq:deepnade:binary:loss} to get a stochastic estimate of the validation set average log-likelihood. An ensemble
    using 16 orderings was used to compute the test-time log-likelihood.
    \end{itemize}

    \begin{table*}
      \small
      \begin{center}
        \caption{Average log-likelihood performance of tractable distribution baselines and NADE models, on binary vector datasets. The best result is shown in bold, along with any other result with an overlapping confidence interval.}
        \smallskip
        \begin{tabular}{lcccccccc}\toprule
          Model       & \footnotesize{Adult} & \footnotesize{Connect4} & \footnotesize{DNA} & \footnotesize{Mushrooms} & \footnotesize{NIPS-0-12} & \footnotesize{OCR-letters} & \footnotesize{RCV1} & \footnotesize{Web} \\
          \midrule
          MoB                  & -20.44      & -23.41      & -98.19      & -14.46      & -290.02      & -40.56          & -47.59               & -30.16 \\
          RBM                  & -16.26      & -22.66      & -96.74      & -15.15      & -277.37      & -43.05          & -48.88               & -29.38 \\
          FVSBN                & \bf{-13.17} & -12.39      & -83.64      & -10.27      & -276.88      & -39.30          & -49.84               & -29.35 \\
          Chow--Liu            & -18.51      & -20.57      & -87.72      & -20.99      & -281.01      & -48.87          & -55.60               & -33.92 \\
          MADE   & \bf{-13.12} & \bf{-11.90} & -83.63      &  -9.68      & -280.25      & -28.34          & -47.10               & -28.53 \\
          MADE-E      & \bf{-13.13} & \bf{-11.90} & \bf{-79.66} &  -9.69      & -277.28      & -30.04          & -46.74               & \bf{-28.25} \\
          \midrule
          NADE   & \bf{-13.19} & -11.99      & -84.81      &  -9.81      & \bf{-273.08} & \bf{-27.22}     & -46.66               & -28.39 \\
          NADE-E      & \bf{-13.19} & -12.58      & -82.31      &  -9.69      & \bf{-272.39} & \bf{-27.32}     & \bf{-46.12}          & \bf{-27.87} \\
          \bottomrule
          \label{tab:bin-results}
        \end{tabular}
      \end{center}
    \end{table*}

    Table~\ref{tab:bin-results} presents the results. We observe that NADE restricted to a fixed ordering of the inputs achieves very competitive performance
    compared to the baselines. However, the order-agnostic version of NADE is overall the best method, being among the top performing model for 5 datasets out of 8.

    The performance of fixed-order NADE is surprisingly robust to variations of the chosen input ordering. The standard deviation on the average log-likelihood when varying the ordering was small: on Mushrooms, DNA and NIPS-0-12, we observed standard deviations of 0.045, 0.05 and 0.15, respectively. However, models with different orders can do well on different test examples, which explains why ensembling can still help.

  \subsection{Binary image dataset}
    \label{sec:results-binimg}

    We now consider the case of an image dataset, constructed by binarizing the MNIST digit dataset, as generated by \citet{Salakhutdinov2008}.
    This benchmark has been a popular choice for the evaluation of generative neural network models.
    Here, we investigate two questions:
    \begin{enumerate}
      \item How does NADE compare to other intractable generative models?
      \item Does the use of a convolutional architecture improve the performance of NADE?
    \end{enumerate}

    For these experiments, in addition to the baselines already described in Section~\ref{sec:binvec}, we consider the following:
    \begin{itemize}
      \item {\bf DARN}~\citep{Gregor2014}: This deep generative autoencoder has two hidden layers, one deterministic and one with binary stochastic units. Both layers have 500 units (denoted as $n_h=500$). Adaptive weight noise (adaNoise) was either used or not to avoid the need for early stopping~\citep{Graves2011}. Evaluation of exact test probabilities is intractable for large latent representations. Hence, Monte Carlo was used to approximate the expected description length, which corresponds to an upper bound on the negative log-likelihood.
      \item {\bf DRAW}~\citep{GregorK2015}: Similar to a variational autoencoder where both the encoder and the decoder are LSTMs, guided (or not) by an attention mechanism. In this model, both LSTMs (encoder and decoder) are composed of 256 recurrent hidden units and always perform 64 timesteps. When the attention mechanism is enabled, patches ($2\times2$ pixels) are provided as inputs to the encoder instead of the whole image and the decoder also produces patches ($5\times5$ pixels) instead of a whole image.
      \item {\bf Pixel RNN}~\citep{OordA2016}: NADE-like model for natural images that is based on convolutional and LSTM hidden units. This model has 7 hidden layers, each composed of 16 units. \cite{OordA2016} proposed a novel two-dimensional LSTM, named Diagonal BiLSTM, which is used in this model. Unlike our ConvNADE, the ordering is fixed before training and at test time, and corresponds to a scan of the image in a diagonal fashion starting from a corner at the top and reaching the opposite corner at the bottom.
    \end{itemize}

    \begin{figure}
      \centering
      \includegraphics[width=0.45\textwidth]{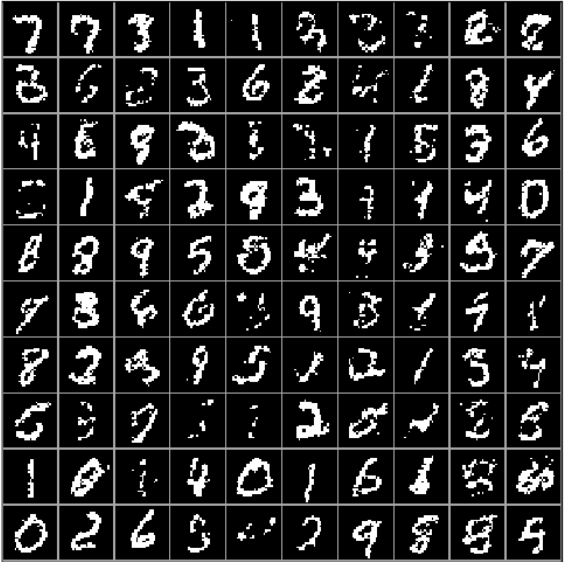}
      \includegraphics[width=0.45\textwidth]{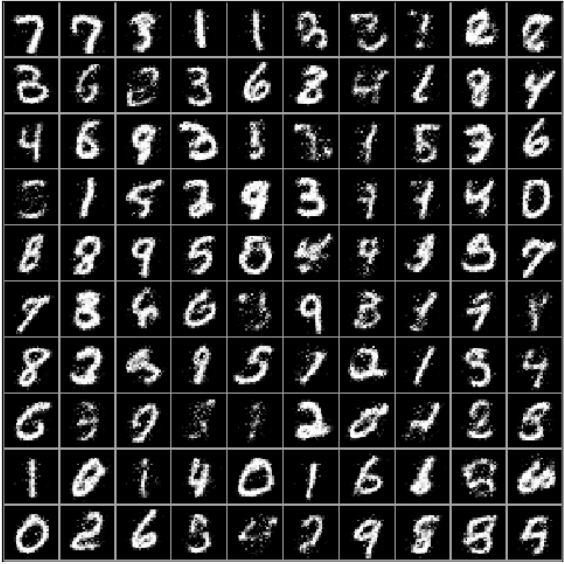}
      \caption{
        {\bf(Left)}: samples from NADE trained on binarized MNIST\@. {\bf(Right)}: probabilities from which each pixel was sampled. Ancestral sampling was used with the same fixed ordering used during training.
      }
      \label{fig:MNIST-nade-samples}
    \end{figure}

    \begin{algorithm}[tb]
      \caption{Pretraining of a NADE with $n$ hidden layers on dataset $X$.}
      \label{alg:deepnade-pretraining}
      \begin{algorithmic}
        \Procedure{PRETRAIN}{$n,X$}
          \If{$n=1$}
            \State \Return \Call{RANDOM-ONE-HIDDEN-LAYER-NADE}{}
          \Else
            \State $nade \gets$ \Call{PRETRAIN}{$n-1, X$}
            \State $nade \gets$ \Call{REMOVE-OUTPUT-LAYER}{$nade$}
            \State $nade \gets$ \Call{ADD-A-NEW-HIDDEN-LAYER}{$nade$}
            \State $nade \gets$ \Call{ADD-A-NEW-OUTPUT-LAYER}{$nade$}
            \State $nade \gets$ \Call{TRAIN-ALL}{$nade,X,iters=20$} \Comment{Train for 20 iterations.}
            \State \Return $nade$
          \EndIf
        \EndProcedure
      \end{algorithmic}
    \end{algorithm}

    \begin{figure}
      \centering
      \subfigure[ConvNADE]{
        \label{fig:convnade-architecture}
        \includegraphics[width=0.3\textwidth]{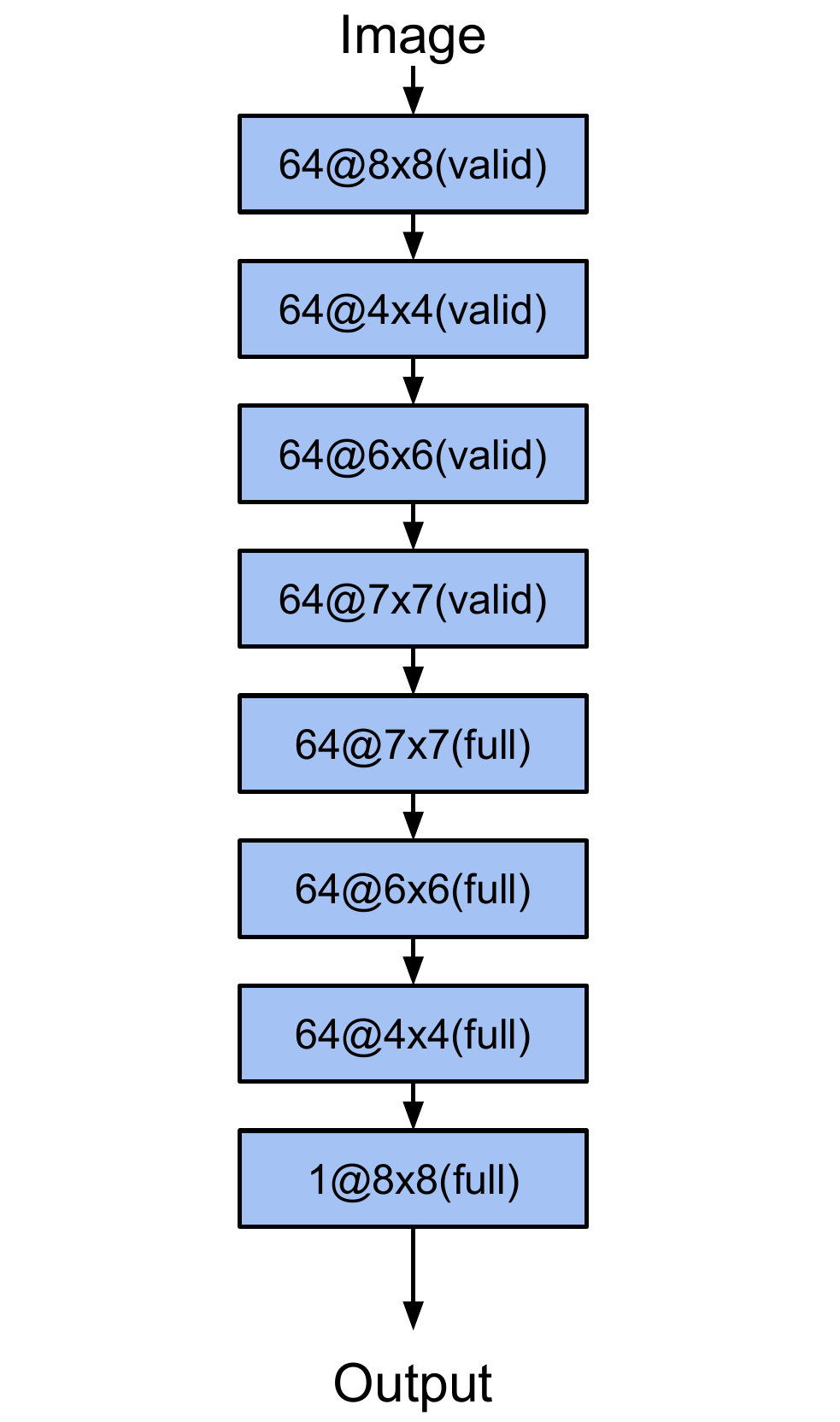}
      }
      \subfigure[ConvNADE + DeepNADE]{
        \label{fig:convnade-deepnade-architecture}
        \includegraphics[width=0.3\textwidth]{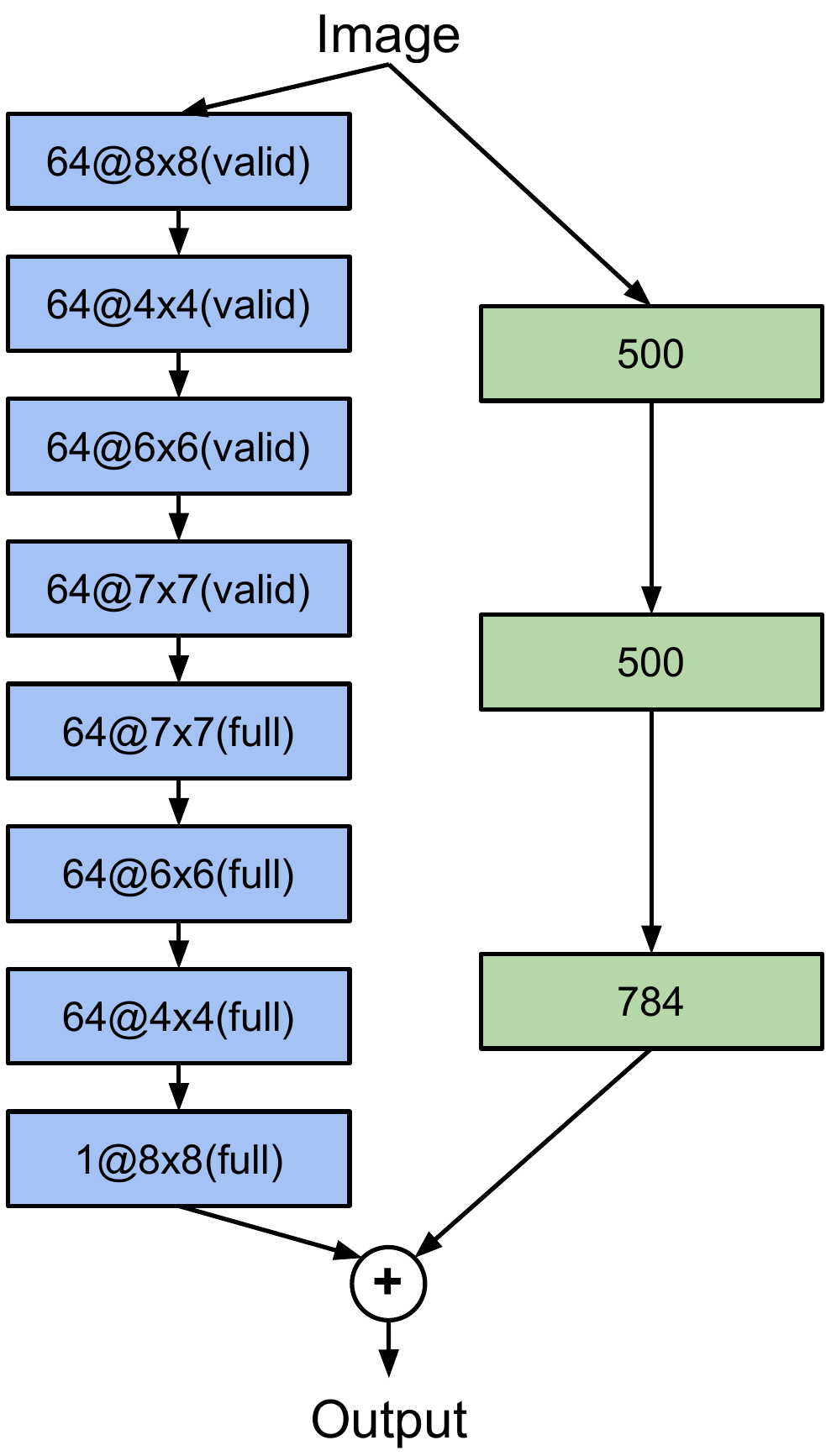}
      }
      \caption{
        Network architectures for binarized MNIST\@. (a) ConvNADE with 8 convolutional layers (depicted in blue). The number of feature maps for a given layer is given by the number before the ``@'' symbol followed by the filter size and the type of convolution is specified in parentheses. (b) The same ConvNADE combined with a DeepNADE consisting of three fully-connected layers of respectively 500, 500 and 784 units.
      }
      \label{fig:architecture-mnist}
    \end{figure}

    We compare these baselines with some NADE variants. The performance of a basic (fixed-order, single hidden layer) NADE model is provided in Table~\ref{tab:deepnade-mnist-results} and samples are illustrated in Figure~\ref{fig:MNIST-nade-samples}. More importantly, we will focus on whether the following variants achieve better test set performance:
    \begin{itemize}
      \item {\bf DeepNADE}: Multiple layers (1hl, 2hl, 3hl or 4hl) trained according to the order-agnostic procedure described in Section~\ref{sec:DeepNADE}. Information about which inputs are masked was either provided or not (no input masks) to the model. The rectified linear activation function was used for all hidden layers. Minibatch gradient descent was used for training, with minibatches of size 1000. Training consisted of 200 iterations of 1000 parameter updates. Each hidden layer was pretrained according to Algorithm~\ref{alg:deepnade-pretraining}. We report an average of the average test log-likelihoods over ten different random orderings.
      \item {\bf EoNADE}: This variant is similar to DeepNADE except for the log-likelihood on the test set, which is instead computed from an ensemble that averages predictive probabilities over 2 or 128 orderings. To clarify, the DeepNADE results report the typical performance of one ordering, by averaging results after taking the log, and so do not combine the predictions of the models like EoNADE does.
      \item {\bf ConvNADE}: Multiple convolutional layers trained according to the order-agnostic procedure described in Section~\ref{sec:DeepNADE}. The exact architecture is shown in Figure~\ref{fig:convnade-architecture}. Information about which inputs are masked was either provided or not (no input masks). The rectified linear activation function was used for all hidden layers. The Adam optimizer~\citep{Kingma2015} was used with a learning rate of $10^{-4}$. Early stopping was used with a look ahead of 10 epochs, using Equation~\ref{eq:deepnade:binary:loss} to get a stochastic estimate of the validation set average log-likelihood. An ensemble using 128 orderings was used to compute the log-likelihood on the test set.
      \item {\bf ConvNADE + DeepNADE}: This variant is similar to ConvNADE except for the aggregation of a separate DeepNADE model at the end of the network. The exact architecture is shown in Figure~\ref{fig:convnade-deepnade-architecture}. The training procedure is the same as with ConvNADE\@.
    \end{itemize}

\begin{figure}
        \centerline{\includegraphics[width=0.5\columnwidth]{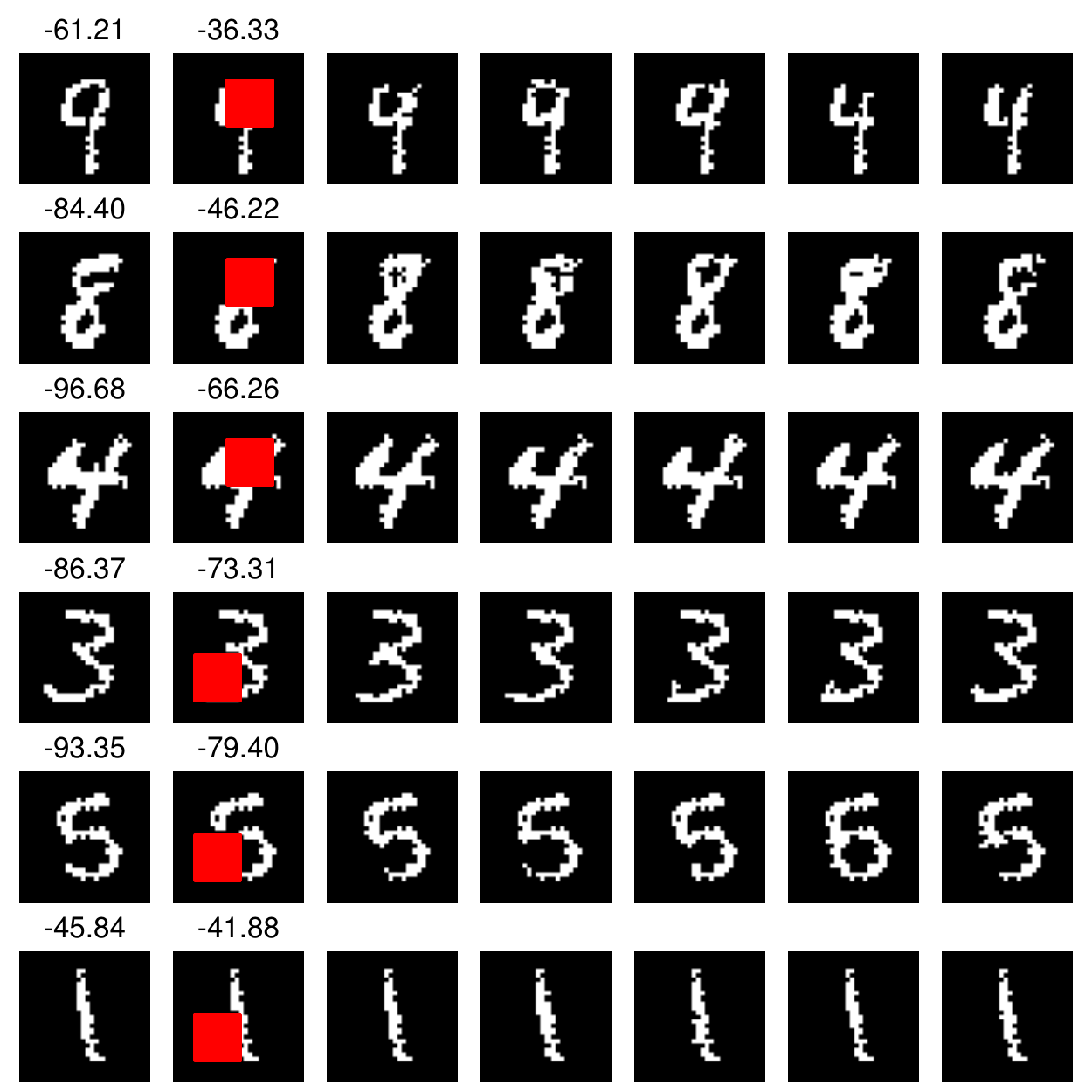}}
        \caption{Example of marginalization and sampling. The first column shows five
            examples from the test set of the MNIST dataset. The second column shows
            the density of these examples when a random 10$\times$10 pixel region is
            marginalized. The right-most five columns show samples for the hollowed
            region. Both tasks can be done easily with a NADE where the pixels to marginalize are at the end of the ordering.}
        \label{fig:mnist-inpainting}
    \end{figure}

    Table~\ref{tab:deepnade-mnist-results} presents the results obtained by models ignorant of the 2D topology, such as the basic NADE model. Addressing the first question, we observe that the order-agnostic version of NADE with two hidden layers is competitive with intractable generative models. Moreover, examples of the ability of DeepNADE to solve inference tasks by marginalization and conditional sampling are shown in Figure~\ref{fig:mnist-inpainting}.

    \begin{table}
      \small
      \begin{center}
        \caption{Negative log-likelihood test results of models ignorant of the 2D topology on the binarized MNIST dataset.}
        \smallskip
        \begin{tabular}{lcc}\toprule
          Model                         & $-\log p$ & $\approx$ \\
          \midrule
          MoBernoullis K=10             & 168.95    &           \\
          MoBernoullis K=500            & 137.64    &           \\
          Chow--Liu tree                & 134.99    &           \\
          MADE 2hl (32 masks)           & 86.64     &           \\
          RBM (500 h, 25 CD steps)      &           & 86.34     \\
          DBN 2hl                       &           & 84.55     \\
          DARN $n_h=500$                &           & 84.71     \\
          DARN $n_h=500$ (adaNoise)     &           & \bf 84.13 \\
          NADE (fixed order)            & 88.33     &           \\
          \midrule
          DeepNADE 1hl (no input masks) & 99.37     &           \\
          DeepNADE 2hl (no input masks) & 95.33     &           \\
          DeepNADE 1hl                  & 92.17     &           \\
          DeepNADE 2hl                  & 89.17     &           \\
          DeepNADE 3hl                  & 89.38     &           \\
          DeepNADE 4hl                  & 89.60     &           \\
          \midrule
          EoNADE 1hl (2 orderings)      & 90.69     &           \\
          EoNADE 1hl (128 orderings)    & 87.71     &           \\
          EoNADE 2hl (2 orderings)      & 87.96     &           \\
          EoNADE 2hl (128 orderings)    & \bf 85.10 &           \\
          \bottomrule
          \label{tab:deepnade-mnist-results}
        \end{tabular}
      \end{center}
    \end{table}

\begin{figure}
      \centering
      \includegraphics[width=0.45\textwidth]{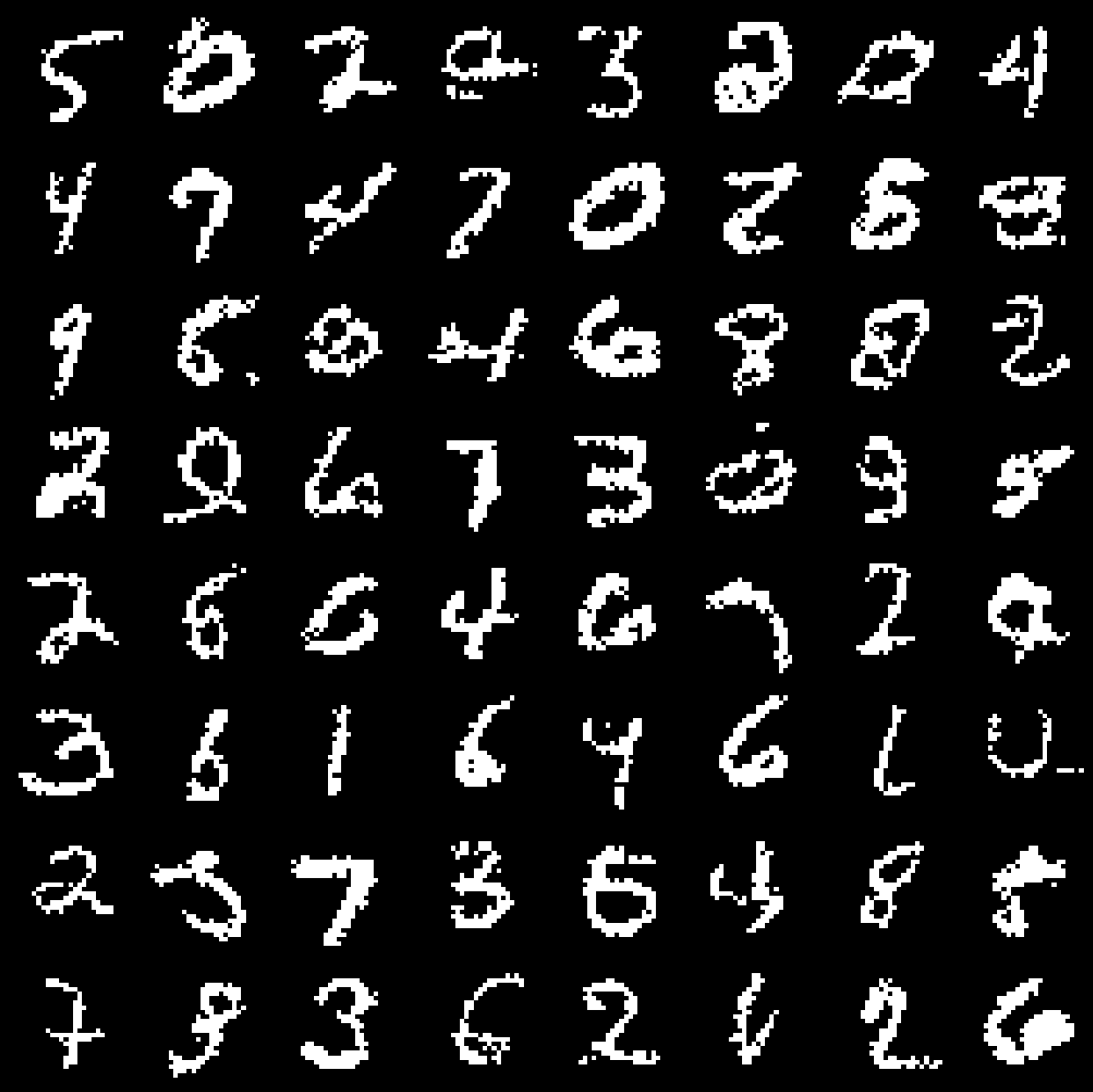}
      \includegraphics[width=0.45\textwidth]{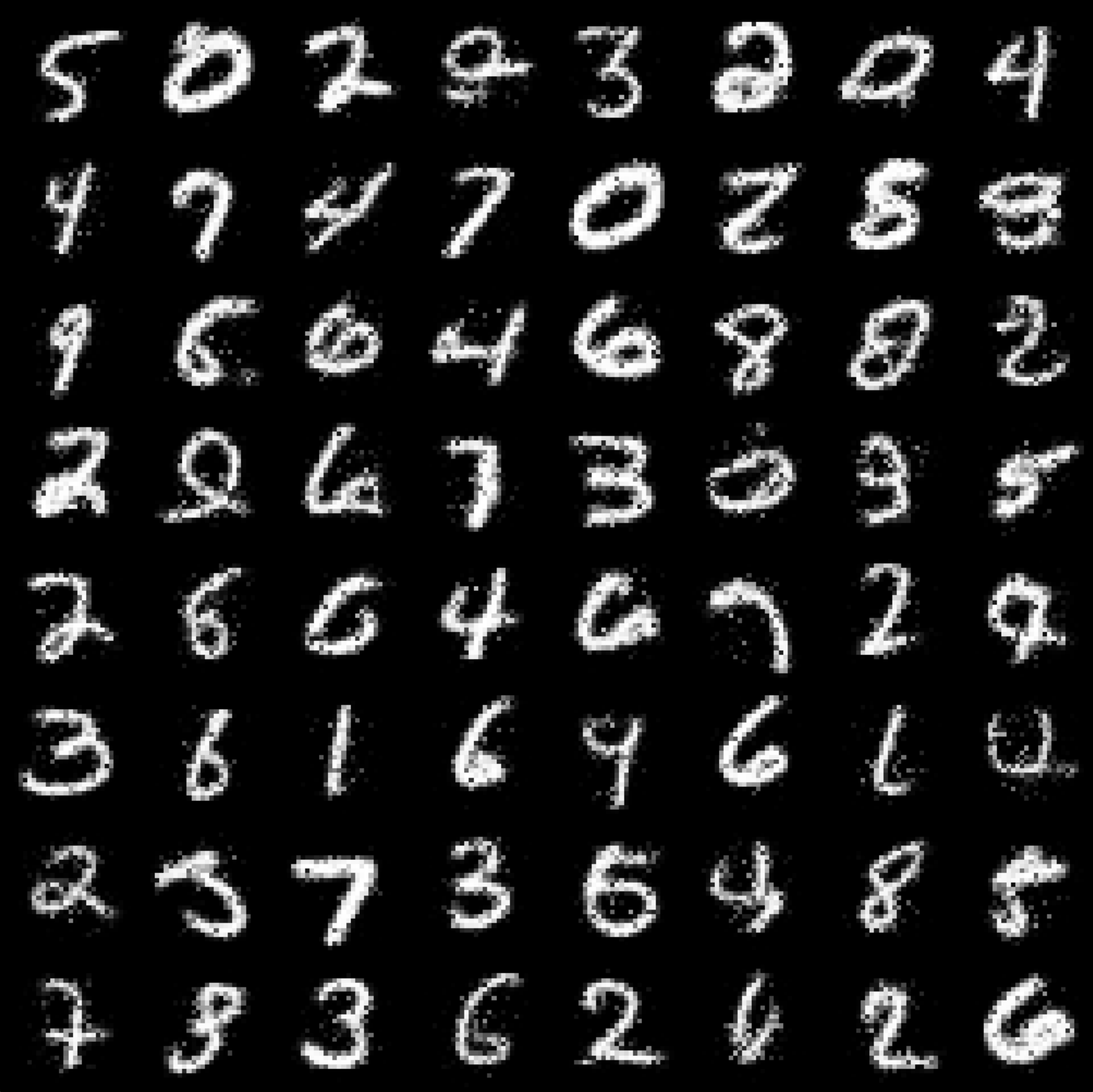}
      \caption{
        {\bf(Left)}: samples from ConvNADE+DeepNADE trained on binarized MNIST\@. {\bf(Right)}: probabilities from which each pixel was sampled. Ancestral sampling was used with a different random ordering for each sample.
      }
      \label{fig:convnade-samples}
    \end{figure}

    Now, addressing the second question, we can see from Table~\ref{tab:mnist-2d-results} that convolutions do improve the performance of NADE\@. Moreover, we observe that providing information about which inputs are masked is essential to obtaining good results. We can also see that combining convolutional and fully-connected layers helps. Even though ConvNADE+DeepNADE performs slightly worst than Pixel RNN, we note that our proposed approach is order-agnostic, whereas Pixel RNN requires a fixed ordering. Figure~\ref{fig:convnade-samples} shows samples obtained from the ConvNADE+DeepNADE model using ancestral sampling on a random ordering.

    \begin{table}
      \small
      \begin{center}
        \caption{Negative log-likelihood test results of models exploiting 2D topology on the binarized MNIST dataset.}
        \smallskip
        \begin{tabular}{lcc}\toprule
          Model       & $-\log p$ & $\le$ \\
          \midrule
          DRAW (without attention)             &           & 87.40 \\
          DRAW                                 &           & 80.97 \\
          Pixel RNN                            & \bf 79.20 &       \\
          \midrule
          ConvNADE+DeepNADE (no input masks)   & 85.25     &       \\
          ConvNADE                             & 81.30     &       \\
          ConvNADE+DeepNADE                    & 80.82     &       \\
          \bottomrule
          \label{tab:mnist-2d-results}
        \end{tabular}
      \end{center}
    \end{table}

\subsection{Real-valued observations datasets}
  
In this section, we compare the statistical performance of RNADE to mixtures of Gaussians~(MoG) and factor analyzers~(MFA), which are surprisingly strong baselines in some tasks~\citep{Tang2012,Zoran2012}.

\subsubsection{Low-dimensional data}
\label{sec:rnade-uci}
We start by considering three UCI datasets~\citep{Bache2013}, previously
used to study the performance of other density
estimators~\citep{Silva2011,Tang2012}, namely: \emph{red wine}, \emph{white wine} and \emph{parkinsons}. These are low dimensional datasets (see
Table~\ref{tab:UCI-characteristics}) with hard thresholds and
non-linear dependencies that make it difficult to fit mixtures of Gaussians or  factor analyzers.

Following~\citet{Tang2012}, we eliminated discrete-valued attributes and an
attribute from every pair with a Pearson correlation coefficient greater
than~0.98. We normalized each dimension of the data by subtracting its training-subset sample mean and dividing by its standard deviation. All results are reported on the normalized data.

We use full-covariance Gaussians and mixtures of factor analysers as baselines.
Models were compared on their log-likelihood on held-out test data. Due to the small size of the datasets (see Table~\ref{tab:UCI-characteristics}), we used
10-folds, using 90\% of the data for training, and 10\% for testing.

We chose the hyperparameter values for each model by doing per-fold
cross-validation, using a ninth of the training data as validation data. Once
the hyperparameter values have been chosen, we train each model using all the
training data (including the validation data) and measure its performance on
the 10\% of held-out testing data. In order to avoid overfitting, we stopped the
training after reaching a training likelihood higher than the one obtained on
the best validation-wise iteration of the best validation run. Early stopping
was important to avoid overfitting the RNADE models. It also improved the
results of the MFAs, but to a lesser degree.

The MFA models were trained using the EM
algorithm~\citep{Ghahramani1996,verbeek2005}. We cross-validated the number of
components and factors. We also selected the number of factors from $2,4,\dots D$, where choosing $D$ results in a mixture of Gaussians, and the number of components was chosen among  $2, 4, \dots 50$. Cross-validation selected fewer than $50$ components in every case.

We report the performance of several RNADE models using different parametric
forms for the one-dimensional conditionals: Gaussian with fixed variance
(RNADE-FV), Gaussian with variable variance (RNADE-Gaussian), $sinh$-$arcsinh$
distribution (RNADE-SAS), mixture of Gaussians (RNADE-MoG), and mixture
of Laplace distributions (RNADE-MoL)\@. All RNADE models were trained by
stochastic gradient descent, using minibatches of size 100, for 500 epochs, each
epoch comprising 10 minibatches. We fixed the number of hidden units to $50$,
and the non-linear activation function of the hidden units to ReLU\@. Three
hyperparameters were cross-validated using grid-search: the number of components
on each one-dimensional conditional (only applicable to the RNADE-MoG and
RNADE-MoL models) was chosen from $\{2,5,10,20\}$, the weight-decay
(used only to regularize the input to hidden weights) from $\left\{2.0,
1.0, 0.1, 0.01, 0.001, 0\right\}$, and the learning rate from $\left\{
0.1, 0.05, 0.025, 0.0125\right\}$. Learning rates were decreased linearly to
reach 0 after the last epoch.

\begin{table}
    \begin{center}
        \caption{Dimensionality and size of the UCI datasets used in Section~\ref{sec:rnade-uci}}
        \label{tab:UCI-characteristics}
        \medskip
        \begin{tabular}{lcccccccc}
            \toprule
            {} & {Red wine} & {White wine} & {Parkinsons} \\
            \midrule
            {Dimensionality} & $11$ & $11$ & $15$ \\
            {Total number of datapoints} & $1599$ & $4898$ & $5875$ \\
            \bottomrule
        \end{tabular}

    \end{center}
\end{table}

\begin{table}
    \begin{center}

        \caption{Average test set log-likelihoods per datapoint for seven models on three UCI datasets. Performances not in bold can be shown to be significantly worse than at least one of the results in bold as per a paired $t$-test on the ten mean-likelihoods (obtained from each data fold), with significance level 0.05.}

        \label{tab:UCI-results}
        \medskip
        \begin{tabular}{lcccccccc}
            \toprule
            {Model} & {Red wine} & {White wine} & {Parkinsons} \\
            \midrule
            {Gaussian} & $-13.18$ & $-13.20$ & $-10.85$ \\
            {MFA} & $-10.19$ & $-10.73$ & $-1.99$ \\
            \hline
            {RNADE-FV}
            &$-12.29$
            &$-12.50$
            &$-8.87$
            \\
            {RNADE-Gaussian}
            &$-11.99$
            &$-12.20$
            &$-3.47$
            \\
            {RNADE-SAS}
            &$-9.86$
            &$-11.22$
            &$-3.07$
            \\
            {RNADE-MoG}
            &$\mathbf{-9.36}$
            &$\mathbf{-10.23}$
            &$\mathbf{-0.90}$
            \\
            {RNADE-MoL}
            &$\mathbf{-9.46}$
            &$-10.38$
            &$-2.63$
            \\
            \bottomrule
        \end{tabular}

    \end{center}
\end{table}

\begin{figure}
    \centerline{\includegraphics[width=1.03\textwidth]{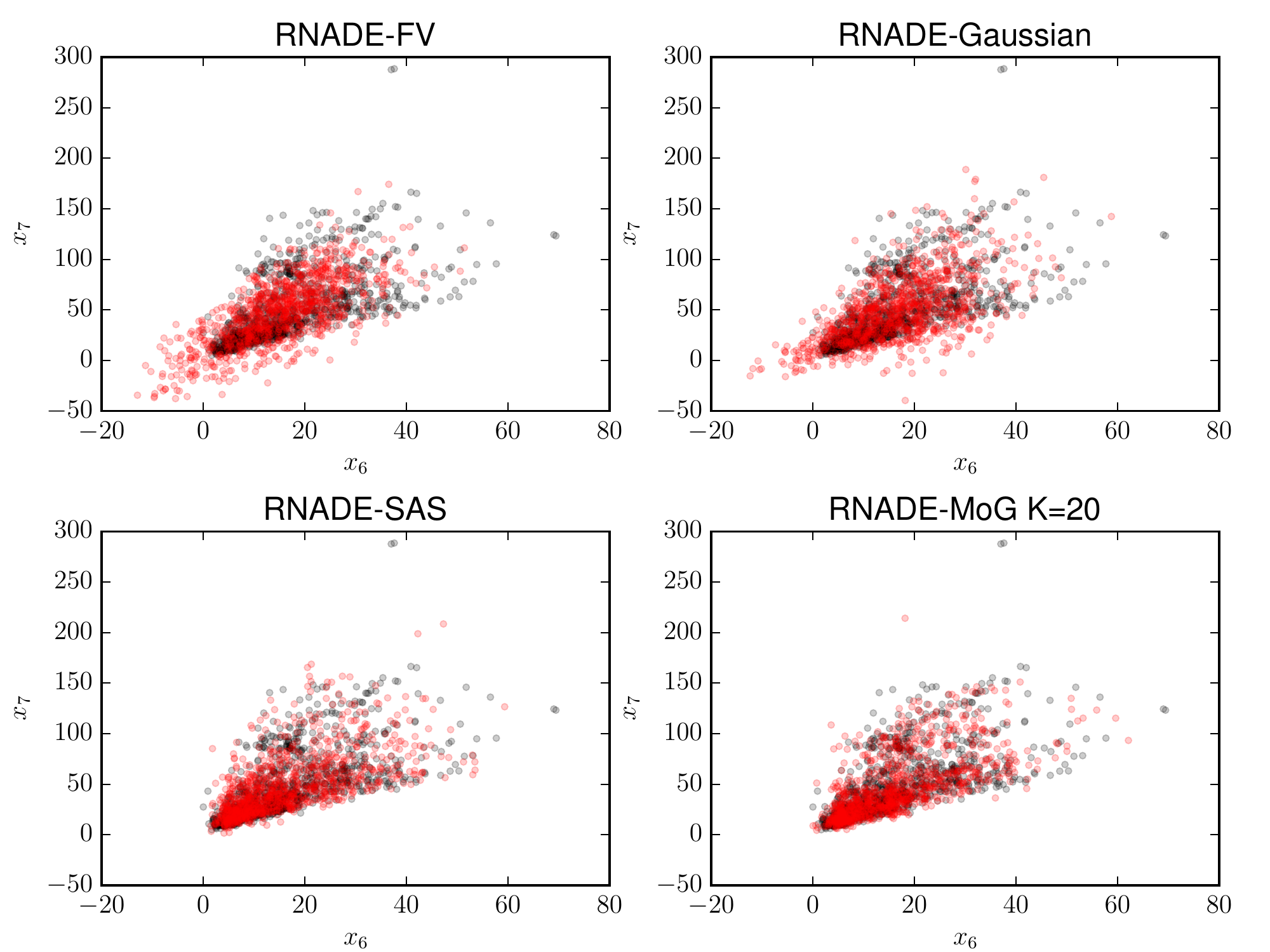}}
    \caption{Scatter plot of dimensions $x_7$ vs $x_6$ of the \textit{red wine} dataset. A thousand datapoints from the dataset are shown in black in all subfigures. As can be observed, this conditional distribution $p(x_7 \given x_6)$ is heteroscedastic, skewed and has hard thresholds. In red, a thousand samples from four RNADE models with different one-dimensional conditional forms are shown. \textbf{Top-left}: In red, one thousand samples from a RNADE-FV model. \textbf{Top-right}: In red, one thousand samples from a RNADE-Gaussian model. \textbf{Bottom-left}: In red, one thousand samples from a RNADE-SAS (sinh-arcsinh distribution) model. \textbf{Bottom-right}: In red, one thousand samples from a RNADE-MoG model with 20 components per one-dimensional conditional.
    The RNADE-SAS and RNADE-MoG models successfully capture all the characteristics of the data.}
    \label{fig:UCI-thresholds}
\end{figure}

The results are shown in Table~\ref{tab:UCI-results}. RNADE with mixture of Gaussian conditionals was among the statistically significant group of best models on all datasets. As shown in Figure~\ref{fig:UCI-thresholds}, RNADE-SAS and RNADE-MoG models are able to capture hard thresholds and heteroscedasticity.

\subsubsection{Natural image patches}
\label{sec:natural-image-patches}
\begin{figure}
    \centerline{\includegraphics[width=1.0\textwidth]{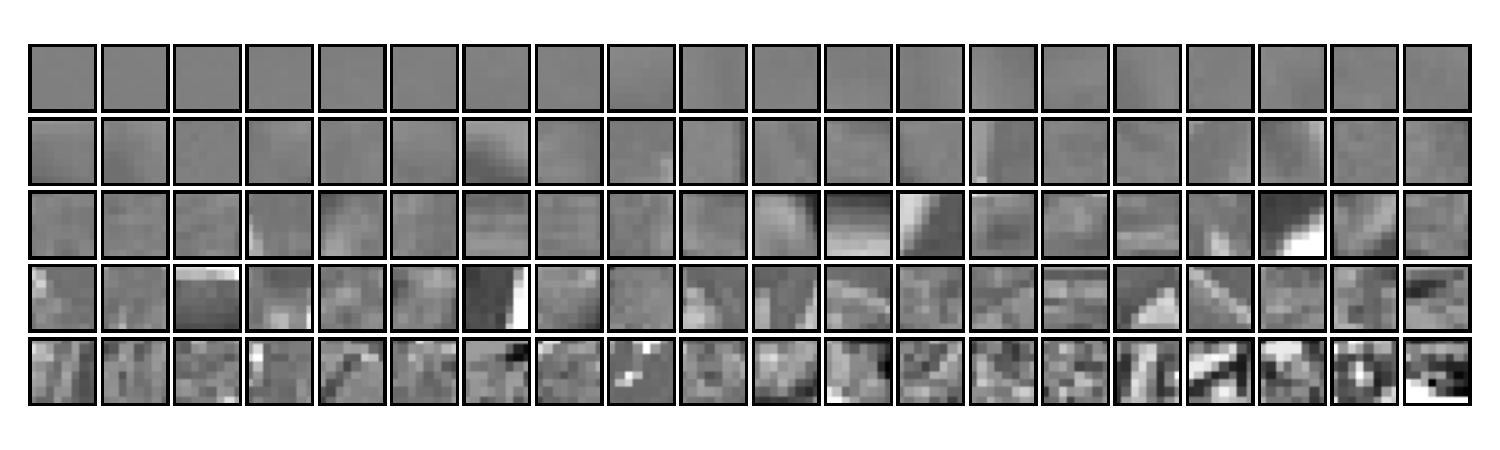}}
    \centerline{\includegraphics[width=1.0\textwidth]{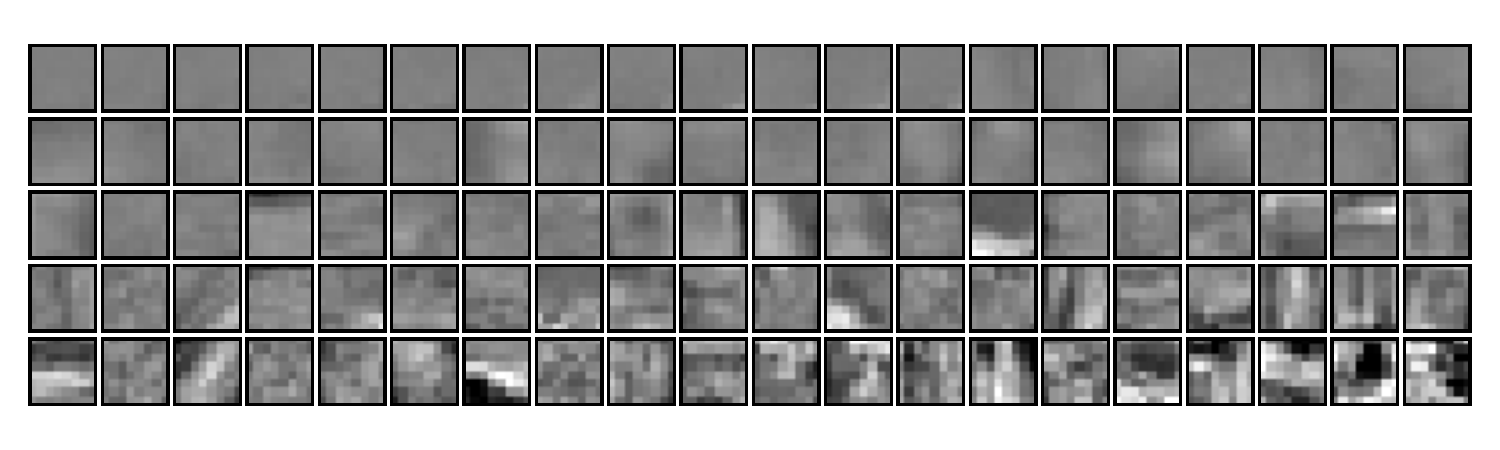}}
    \centerline{\includegraphics[width=1.0\textwidth]{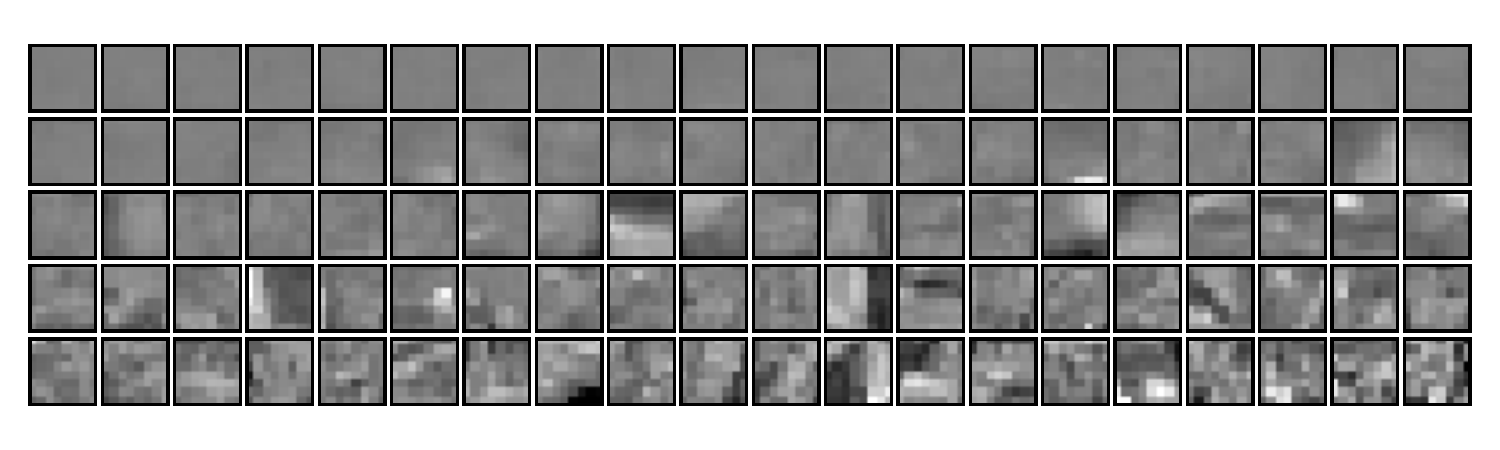}}
    \caption{
        \textbf{Top:}~100 8$\times$8 patches from the BSDS test set.
        \textbf{Center:}~100 samples from
        a mixture of Gaussians with 200 full-covariance components. \textbf{Bottom:}~100 samples from
        an RNADE with 1024 hidden units and 10 Gaussian components per conditional. All data and samples were drawn randomly and sorted by their density under the RNADE\@.}
    \label{fig:natimg-samples}
\end{figure}

We also measured the ability of RNADE to
model small patches of natural images.  Following the work of \citet{Zoran2011}, we use 8-by-8-pixel patches of monochrome natural images, obtained from the BSDS300 dataset~\citetext{\citealp{Martin2001}; Figure~\ref{fig:natimg-samples} gives examples}.

Pixels in this dataset can take a finite number of brightness values ranging
from $0$ to~$255$. We added uniformly distributed noise between
$0$ and~$1$ to the brightness of each pixel. We then divided by~$256$, making
the pixels take continuous values in the range~$[0,1]$. Adding noise prevents deceivingly high-likelihood solutions that assign narrow high-density spikes around some of the possible discrete values.

We subtracted the mean pixel value from each patch. Effectively reducing the dimensionality of the data. Therefore we discarded the 64th (bottom-right) pixel, which would be perfectly predictable and models could fit arbitrarily high densities to it. All of the results in this section were obtained by fitting the pixels in a
raster-scan order.

Experimental details follow. We trained our models by using patches randomly
drawn from 180 images in the training subset of BSDS300. We used the remaining
20 images in the training subset as validation data. We used 1000 random patches
from the validation subset to early-stop training of RNADE\@. We measured the
performance of each model by their log-likelihood on one million patches drawn
randomly from the test subset of 100 images not present in the training data.
Given the larger scale of this dataset, hyperparameters of the RNADE and MoG
models were chosen manually using the performance of preliminary runs on the
validation data, rather than by grid search.

All RNADE models reported use ReLU activations for the hidden units. The RNADE models were trained by stochastic gradient descent, using 25
datapoints per minibatch, for a total of 1,000 epochs, each comprising 1,000
minibatches. The learning rate was initialized to 0.001, and linearly decreased to reach 0 after the last epoch. Gradient momentum with factor 0.9 was
used, but initiated after the first epoch. A weight decay rate of
0.001 was applied to the input-to-hidden weight matrix only. We found that
multiplying the gradient of the mean output parameters by the standard deviation
improves results of the models with mixture outputs\footnote{Empirically, we
    found this to work better than regular gradients and also better than
    multiplying by the variances, which would provide a step with the right units.}.
RNADE training was early stopped but didn't show signs of overfitting. Even
larger models might perform better.

The MoG models were trained using 1,000 iterations of minibatch EM\@. At each
iteration 20,000 randomly sampled datapoints were used in an EM update. A step
was taken from the previous parameters' value towards the parameters resulting from the M-step: $\vec{\theta_{t}} =
(1-\eta)\vec{\theta_{t-1}}+\eta\vec{\theta_{EM}}$. The step size, $\eta$,
was scheduled to start at 0.1 and linearly decreased to reach~0 after the last
update. The training of the MoG was early-stopped and also showed no signs
of overfitting.

The results are shown in Table~\ref{tab:BSDS-rnade-results}. We report the
average log-likelihood of each model for a million image patches from the test set. The ranking of RNADE models is maintained when ordered by validation
likelihood: the model with best test-likelihood would have been chosen using
crossvalidation across all the RNADE models shown in the table. We also compared
RNADE with a MoG trained by Zoran and Weiss (downloaded from Daniel Zoran's
website) from which we removed the 64th row and column of each covariance
matrix. There are two differences in the set-up of our experiments and those of
Zoran and Weiss. First, we learned the means of the MoG components,
while~\cite{Zoran2011} fixed them to zero. Second, we held-out 20 images from
the training set to do early-stopping and hyperparameter optimisation, while
they used the 200 images for training.

\begin{table}
    \begin{minipage}{\textwidth}
    \begin{center}
        \caption{Average per-example log-likelihood of several mixture of
            Gaussian and RNADE models on 8$\times$8 pixel patches of natural images. These results are reported in nats and were calculated using one million patches. Standard errors due to the finite test sample size are lower than~$0.1$ nats in every case. $h$ indicates the number of hidden units in the RNADE models, and  $K$ the number of one-dimensional components for each conditional in RNADE or the number of full-covariance components for~MoG\@.}
        \label{tab:BSDS-rnade-results}
        \medskip
        \begin{tabular}{lc}
            \toprule
            {Model} & {Test log-likelihood}\\
            \midrule
            MoG $K\!=\!200$ \citep{Zoran2012}\footnote{This model was trained using the full 200 images in the BSDS training dataset, the rest of the models were trained using 180, reserving 20 for hyperparameter crossvalidation and early-stopping.} & 152.8\\
            MoG $K\!=\!100$ & 144.7\\
            MoG $K\!=\!200$ & 150.4 \\
            MoG $K\!=\!300$ & 150.4\\
            \midrule
            RNADE-FV $h\!=\!512$ & 100.3 \\
            RNADE-Gaussian $h\!=\!512$ & 143.9 \\
            RNADE-Laplace $h\!=\!512$ & 145.9 \\
            RNADE-SAS\footnote{Training an RNADE with sinh-arcsinh conditionals required the use of a starting learning rate 20 times smaller to avoid divergence during training. For this reason, this model was trained for 2000 epochs.} $h\!=\!512$ & 148.5 \\
            \midrule
            RNADE-MoG $K\!=\!2$ $h\!=\!512$ & 149.5\\
            RNADE-MoG $K\!=\!2$ $h\!=\!1024$ & 150.3\\
            RNADE-MoG $K\!=\!5$ $h\!=\!512$ & 152.4\\
            RNADE-MoG $K\!=\!5$ $h\!=\!1024$ & 152.7\\
            RNADE-MoG $K\!=\!10$ $h\!=\!512$ & 153.5\\
            RNADE-MoG $K\!=\!10$ $h\!=\!1024$ & \textbf{153.7}\\
            \midrule
            RNADE-MoL $K\!=\!2$ $h\!=\!512$ & 149.3\\
            RNADE-MoL $K\!=\!2$ $h\!=\!1024$ & 150.1\\
            RNADE-MoL $K\!=\!5$ $h\!=\!512$ & 151.5\\
            RNADE-MoL $K\!=\!5$ $h\!=\!1024$ & 151.4\\
            RNADE-MoL $K\!=\!10$ $h\!=\!512$ & 152.3\\
            RNADE-MoL $K\!=\!10$ $h\!=\!1024$ & 152.5\\
            \bottomrule
        \end{tabular}
    \end{center}
    \end{minipage}
\end{table}

The RNADE-FV model with fixed conditional variances obtained
very low statistical performance. Adding an output parameter per dimension to
have variable standard deviations made our models competitive with MoG with 100
full-covariance components. However, in order to obtain results superior to the
mixture of Gaussians model trained by Zoran and Weiss, we had to use richer
conditional distributions: one-dimensional mixtures of Gaussians (RNADE-MoG)\@.
On average, the best RNADE model obtained 3.3\;nats per patch higher log-density
than a MoG fitted with the same training data.

In Figure~\ref{fig:natimg-samples}, we show one hundred examples from the test set, one hundred examples from Zoran and Weiss' mixture of Gaussians, and a hundred samples from our best RNADE-MoG model. Similar patterns can be observed in the three cases: uniform patches, edges, and locally smooth noisy patches.

\subsubsection{Speech acoustics}

We also measured the ability of RNADE to model small patches of speech
spectrograms, extracted from the TIMIT dataset~\citep{Garofolo1993}. The patches
contained 11 frames of 20 filter-banks plus energy; totalling 231 dimensions per
datapoint.
A good generative model of speech acoustics could be used, for
example, in denoising, or speech detection tasks.

We fitted the models using the standard TIMIT training subset, which includes
recordings from 605 speakers of American English. We compare RNADE with a
mixture of Gaussians by measuring their log-likelihood on the complete TIMIT
core-test dataset: a held-out set of 25 speakers.

The RNADE models have 512 hidden units, ReLU activations, and a mixture of 20
one-dimensional Gaussian components per output. Given the large scale of this
dataset, hyperparameter choices were again made manually using validation data. The same minibatch training procedures for RNADE and mixture of Gaussians were used as for natural image patches.

The RNADE models were trained by stochastic gradient descent, with 25 datapoints
per minibatch, for a total of 200 epochs, each comprising 1,000 minibatches. The
learning rate was initialized to 0.001 and linearly decreased to reach 0
after the last epoch. Gradient momentum with momentum factor 0.9 was used, but
initiated after the first epoch. A weight decay rate of 0.001 was
applied to the input-to-hidden weight matrix only. Again, we found that
multiplying the gradient of the mean output parameters by the standard deviation
improved results. RNADE training was early stopped but didn't show signs of
overfitting.

As for the MoG model, it was trained exactly as in Section~\ref{sec:natural-image-patches}.

\begin{figure}[!t]
    \includegraphics[width=1\textwidth]{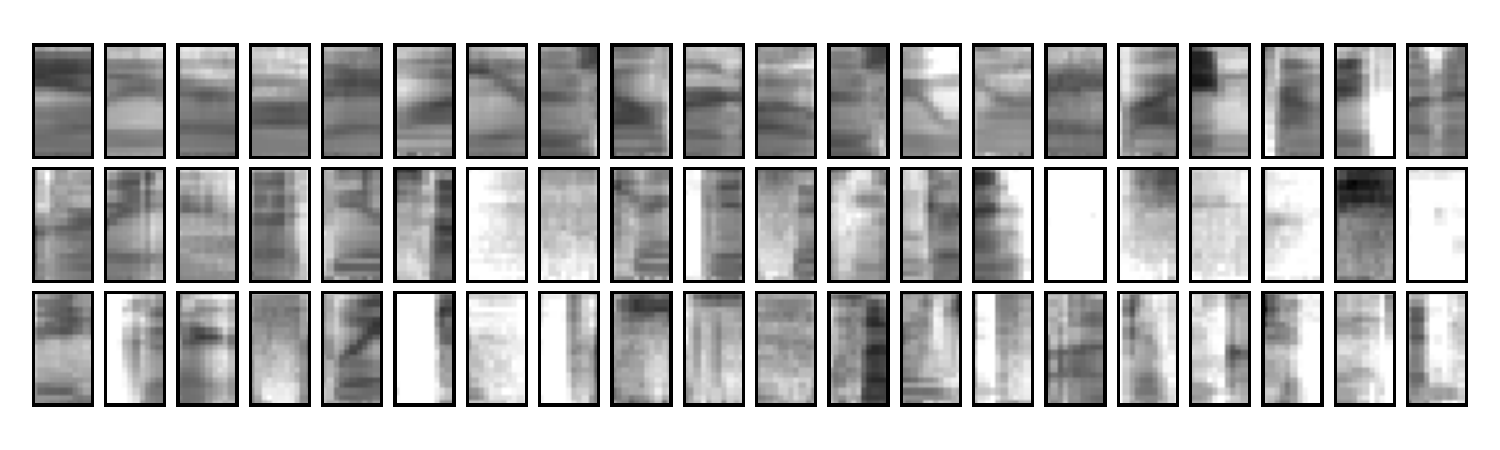}
    \includegraphics[width=1\textwidth]{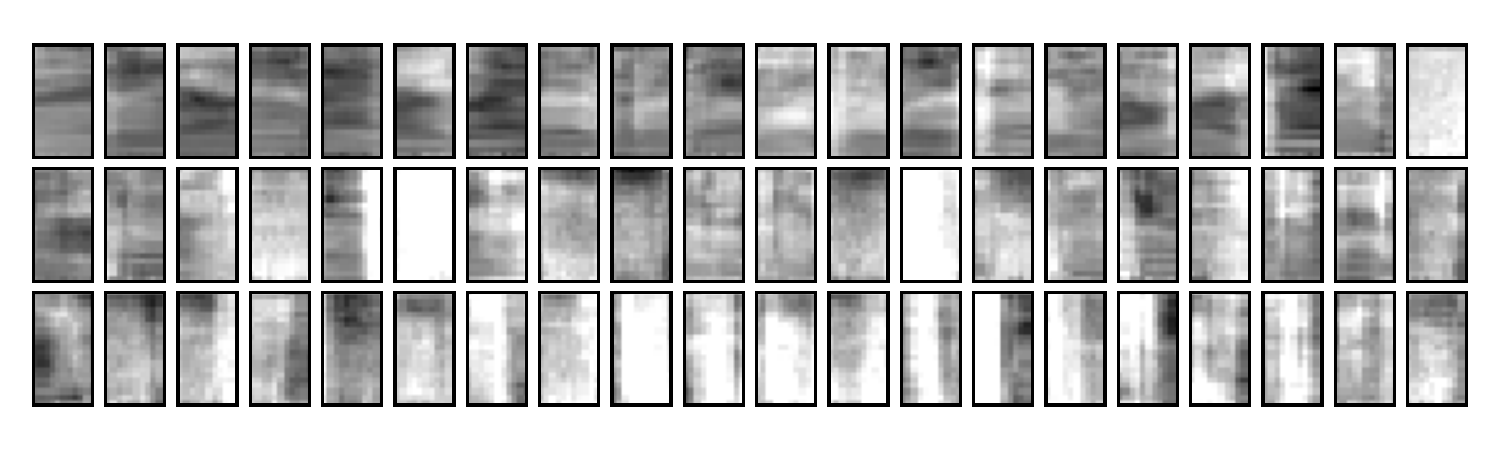}
    \includegraphics[width=1\textwidth]{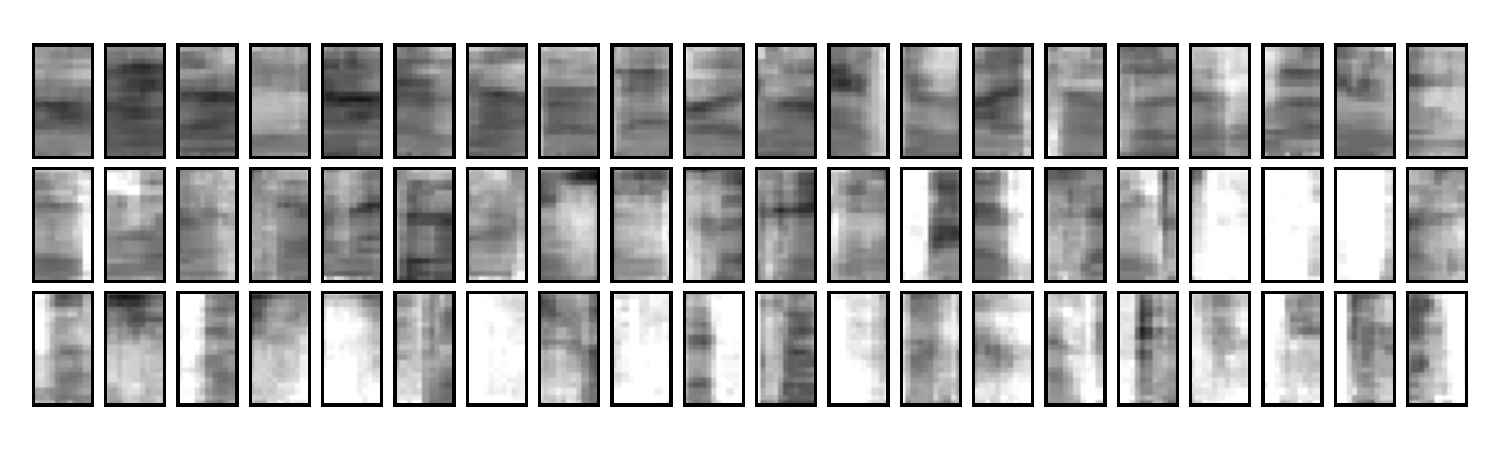}

    \caption{\textbf{Top:}~60 datapoints from the TIMIT core-test set. \textbf{Center:}~60 samples from a MoG model with 200 components. \textbf{Bottom:}~60 samples from an RNADE with 10 Gaussian output components per dimension. For each datapoint displayed, time is shown on the horizontal axis, the bottom row displays the energy feature, while the others display the Mel filter bank features (in ascending frequency order from the bottom). All data and samples were drawn randomly and sorted by density under the RNADE model.}

    \label{fig:speech-samples}
\end{figure}

The results are shown in Table \ref{tab:TIMIT-results}. The best RNADE (which
would have been selected based on validation results) has 15 nats higher
likelihood per test example than the best mixture of Gaussians. Examples from
the test set, and samples from the MoG and RNADE-MoG models are shown in
Figure~\ref{fig:speech-samples}. In contrast with the log-likelihood measure,
there are no marked differences between the samples from each model. Both sets of
samples look like blurred spectrograms, but RNADE seems to capture sharper
formant structures (peaks of energy at the lower frequency bands characteristic
of vowel sounds).

\begin{table}[!t]
    \begin{center}
    \caption{Log-likelihood of several MoG and RNADE models on the core-test
        set of TIMIT measured in nats. Standard errors due to the finite test
        sample size are lower than $0.4$ nats in every case. RNADE obtained a
        higher (better) log-likelihood.}
    \label{tab:TIMIT-results}
    \medskip
    \begin{tabular}{lcc}
        \toprule
        {Model} & {Test LogL} \\
        \midrule
        MoG $N\!=\!50$  & 110.4\\
        MoG $N\!=\!100$ & 112.0\\
        MoG $N\!=\!200$ & 112.5\\
        MoG $N\!=\!300$ & 112.5\\
        \midrule
        RNADE-Gaussian & 110.6\\
        RNADE-Laplace & 108.6\\
        RNADE-SAS & 119.2\\
        RNADE-MoG $K\!=\!2$ & 121.1\\
        RNADE-MoG $K\!=\!5$ & 124.3\\
        RNADE-MoG $K\!=\!10$ & \textbf{127.8}\\
        RNADE-MoL $K\!=\!2$ & 116.3\\
        RNADE-MoL $K\!=\!5$ & 120.5\\
        RNADE-MoL $K\!=\!10$ & 123.3\\
        \bottomrule
    \end{tabular}
    \end{center}
\end{table}

\section{Conclusion}
\label{sec:conclusion}

We've described the Neural Autoregressive Distribution Estimator, a tractable, flexible and competitive
alternative to directed and undirected graphical models for unsupervised distribution estimation.

Since the publication of the first formulation of NADE~\citep{Larochelle2011}, it has been
extended to many more settings, other than those described in this paper. \citet{Larochelle2012,ZhengY2015b}
adapted NADE for topic modeling of documents and images, while \citet{Boulanger-LewandowskiN2012} used
NADE for modeling music sequential data. \citet{TheisL2015} and \citet{OordA2016}
proposed different NADE models for images than the one we presented, applied
to natural images and based on convolutional and LSTM hidden units. \citet{ZhengY2015} used a NADE model
to integrate an attention mechanism into an image classifier. \citet{BornscheinJ2015} showed
that NADE could serve as a powerful prior over the latent state of directed graphical model.
These are just a few examples of many possible ways one can leverage the flexibility and effectiveness
of NADE models.

\bibliography{NADE}

\end{document}